\pdfoutput=1
\UseRawInputEncoding
\documentclass[nohyperref]{article}

\usepackage[dvipsnames]{xcolor}

\usepackage{multicol}
\usepackage{microtype}
\usepackage{graphicx}
\usepackage{subfigure}
\usepackage{booktabs} % for professional tables
\usepackage{caption}
\usepackage{hyperref}

\usepackage[accepted]{icml2022}

\usepackage{amsmath}
\usepackage{amssymb}
\usepackage{mathtools}
\usepackage{amsthm}

\usepackage[capitalize,noabbrev]{cleveref}

\theoremstyle{plain}

\theoremstyle{definition}

\theoremstyle{remark}

\usepackage{amsmath,amsfonts,bm}

\def\eqref#1{equation~\ref{#1}}
\def\1{\bm{1}}

\DeclareMathAlphabet{\mathsfit}{\encodingdefault}{\sfdefault}{m}{sl}
\SetMathAlphabet{\mathsfit}{bold}{\encodingdefault}{\sfdefault}{bx}{n}

\usepackage{hyperref}
\usepackage{url}
\usepackage{enumitem}

\usepackage{adjustbox}
\usepackage{hyperref}       % hyperlinks
\usepackage{booktabs}
\usepackage{diagbox}
\usepackage{multirow}

\usepackage{algorithm}
\usepackage{algorithmic}
\usepackage{listings}

\usepackage[textsize=tiny]{todonotes}

\icmltitlerunning{Revisiting Label Smoothing and Knowledge Distillation Compatibility: What was Missing?}

\begin{document}

\twocolumn[
\icmltitle{Revisiting Label Smoothing and Knowledge Distillation Compatibility\\ What was Missing?}



\icmlsetsymbol{equal}{*}

\begin{icmlauthorlist}
\icmlauthor{Keshigeyan Chandrasegaran}{sutd}
\icmlauthor{Ngoc-Trung Tran}{sutd,equal}
\icmlauthor{Yunqing Zhao}{sutd,equal}
\icmlauthor{Ngai-Man Cheung}{sutd}
\end{icmlauthorlist}

\icmlaffiliation{sutd}{Singapore University of Technology and Design (SUTD)}

% \icmlcorrespondingauthor{Keshigeyan Chandrasegaran}{keshigeyan@sutd.edu.sg}
\icmlcorrespondingauthor{Ngai-Man Cheung}{ngaiman\_cheung@sutd.edu.sg}

\icmlkeywords{Machine Learning, ICML}

\vskip 0.3in
]

\printAffiliationsAndNotice{
\icmlEqualContribution
} % otherwise use the standard text

% \printAffiliationsAndNotice{}

% \footnote{{Preliminary work.  Under review by the
% International Conference on Machine Learning (ICML)\@.  Do not distribute.}}

% \let\thefootnote\relax\footnotetext{\hspace*{-\footnotesep}
% {Preliminary work. Under review by the International Conference on Machine Learning (ICML)\@.  Do not distribute.}
% }

\newcommand{\fix}{\marginpar{FIX}}
\newcommand{\new}{\marginpar{NEW}}

\begin{abstract}
This work investigates the compatibility between label smoothing (LS) and knowledge distillation (KD). 
Contemporary findings addressing this thesis statement take dichotomous standpoints: \citet{muller,shen2021}.
Critically, there is no effort to understand and resolve these contradictory findings, 
leaving the primal question $-$ to smooth or not to smooth a teacher network? $-$ unanswered.
The main contributions of our work are the 
discovery, analysis and validation of {\em systematic diffusion} as 
the missing concept which is instrumental in understanding and resolving these contradictory findings. 
This systematic diffusion essentially curtails the benefits of distilling from an LS-trained teacher, thereby rendering KD at increased temperatures ineffective.
Our discovery is comprehensively supported by large-scale experiments, analyses and case studies including image classification, neural machine translation and compact student distillation tasks spanning across multiple datasets and teacher-student architectures. 
Based on our analysis, we suggest practitioners to use an LS-trained
teacher with a low-temperature transfer to
achieve high performance students.
Code and models are available at \url{https://keshik6.github.io/revisiting-ls-kd-compatibility/}
\end{abstract}

\vspace{-0.5cm}
\section{Introduction}

This paper deeply investigates the compatibility between label smoothing \citep{szegedy_ls} and knowledge distillation \citep{hinton_kd}.
Specifically, we aim to 
% explain 
revisit
and resolve the contradictory standpoints of \citet{muller} and \citet{shen2021}, thereby establishing a foundational understanding on the compatibility between label smoothing (LS) and knowledge distillation (KD). 
Both LS and KD involve training a model (i.e.: deep neural networks) with soft-targets. 
In LS, instead of computing cross entropy loss with the hard-target (one-hot encoding) of a training sample, a soft-target is used, which is a weighted mixture of the one-hot encoding and the uniform distribution.
A mixture parameter $\alpha$ is used in LS to specify the extent of mixing.
On the other hand, KD involves training a teacher model (usually a powerful model) and a student model (usually a compact model). 
The objective of KD is to transfer knowledge from the teacher model to the student model. 
In the most common form, the student model is trained to match the soft output of the teacher model. 
The success of KD has been attributed to the transference of logits' information about resemblances between instances of different classes (logits are the inputs to the final softmax which produces the soft targets).
In KD \citep{hinton_kd}, a temperature $T$ is introduced to facilitate the transference: an increased $T$ may produce more suitable soft targets that have more emphasis on the probabilities of incorrect classes (or equivalently, logits of the incorrect classes).

\textbf{LS and KD research dialogue.} 
Recently, a notable amount of research efforts has been conducted to understand the relationship between LS and KD
\citep{muller, shen2021, lukasik20a, Yuan_2020_CVPR, tang2021}. 
One of the most intriguing and controversial discussion is the compatibility between LS and KD. Particularly, \textit{in KD, does label smoothing in a teacher network suppress the effectiveness of the distillation?}

\citet{muller} are the first to investigate this topic, and their findings suggest that applying LS to a teacher network impairs the performance of KD. 
In particular, they visualize the penultimate layer representations in the teacher network to show that LS erases information in the logits about resemblances between instances of different classes. 
Since this information is essential for KD, they conclude that applying LS for the teacher network can hurt KD.
$\bullet$
``If a teacher network is trained with label smoothing, knowledge distillation into a student network is much less effective.'' \citep{muller}  
$\bullet$
``Label smoothing can hurt distillation'' \citep{muller}

The conclusion of \citet{muller} is widely accepted \citep{khosla, Arani_2021_WACV, tang2021, MghabbarR20, shen2021unmix}. However, very recently, this is questioned by \citet{shen2021}. 
In particular, their work discussed a new finding: information erasure in teacher can actually enlarge the central distance between \emph{semantically similar classes}, allowing the student to learn to classify these categories easily.
\citet{shen2021} claim that this benefit of using an LS-trained teacher outweighs the detrimental effect due to information erasure. Therefore, they conclude that LS in a teacher network does not suppress the effectiveness of KD. 
$\bullet$
``Label smoothing will not impair the predictive performance of students.'' \citep{shen2021}  
$\bullet$
``Label smoothing is compatible with knowledge distillation'' \citep{shen2021}

\textbf{LS and KD compatibility remains unresolved.}
We were perplexed by the seemingly contradictory findings by \citet{muller} and \citet{shen2021}. 
While the latter has shown empirical results to support their own finding, their work does not investigate the opposite standpoint and contradictory results by \citet{muller}.
\textit{
Critically, there is no effort to understand and resolve the seemingly contradictory arguments and supporting evidences by \citet{muller} and \citet{shen2021}.
}
Consequently, for practitioners, it remains unclear as to under what situations LS can be applied to the teacher network in KD, and under what situations it must be avoided.

% Our contributions
\textbf{Our contributions.}
We begin by meticulously scrutinizing the opposing findings of \citet{muller} and \citet{shen2021}. 
In particular, we discover that in the presence of an LS-trained teacher, KD at higher temperatures \emph{systematically} diffuses penultimate layer representations learnt by the student towards semantically similar classes.  This systematic diffusion essentially curtails the benefits (as claimed by \citet{shen2021}) obtained by distilling from an LS-trained teacher, thereby rendering KD at increased temperatures ineffective.
We perform large-scale KD experiments including image classification using ImageNet-1K \cite{imagenet_cvpr09}, fine-grained image classification using CUB200-2011 \cite{WahCUB_200_2011}, 
neural machine translation (English $\rightarrow$ German, English $\rightarrow$ Russian translation) using IWSLT, 
compact student distillation (MobileNetV2 \cite{sandler2018mobilenetv2}, EfficientNet-B0 \citep{tan2019efficientnet}) and 
multiple teacher-student architectures
to comprehensively demonstrate this systematic diffusion in the student qualitatively using penultimate layer visualizations, and quantitatively using our proposed relative distance metric called diffusion index ($\eta$). 

Our finding on \emph{systematic} diffusion is very critical when distilling from an LS-trained teacher.
Particularly, we argue that this \textit{diffusion} maneuvers the penultimate layer representations learnt by the student of a given class in a \textit{systematic} way that targets in the direction of semantically similar classes. Therefore, this systematic diffusion directly  curtails the distance enlargement (between semantically similar classes) benefits obtained by distilling from an LS-trained teacher.
Our qualitative and quantitative analysis with our proposed relative distance metric ($\eta$) in Sec. \ref{sec:systematic-diffusion} aims to establish not only the existence of this diffusion, but also establish that such diffusion is \emph{systematic}
and not isotopic.

Importantly, using systematic diffusion analysis, we explain and resolve the contradictory findings by \citet{muller} and \citet{shen2021}, thereby establishing a foundational understanding on the compatibility between LS and KD.
Finally, using our discovery on systematic diffusion, we provide empirical guidelines for practitioners regarding the combined use of LS and KD. 
We summarize our key findings in Table \ref{table:main-findings}.
\textbf{The key takeaway from our work is:}

\vspace{-0.4cm}
\begin{itemize}[leftmargin=*]
    \item
    In the presence of an LS-trained teacher, KD at higher temperatures systematically diffuses penultimate layer representations learnt by the student towards semantically similar classes. 
    This systematic diffusion essentially curtails the benefits of distilling from an LS-trained teacher, thereby rendering KD at increased temperatures ineffective.
    Specifically, systematic diffusion was the missing concept that is instrumental in explaining and resolving the contradictory findings of \citet{muller} and \citet{shen2021}, 
    thereby clearing up the existential conundrum regarding the compatibility between LS and KD.
\end{itemize}
\vspace{-0.1cm}
% End of introduction -----------

%% ---------Show our contribution as table to make it clear --------------
\begin{table*}[h]
    \centering
    \caption{
    Main findings regarding LS and KD compatibility in recent works and our work.
    }
    \begin{adjustbox}{width=0.87\textwidth,center}
        %\begin{tabular}{ll|C{3cm}|l|l|l}\toprule
        \begin{tabular}{l|p{1.2cm}|p{2.6cm}|p{3.0cm}|p{4.2cm}|p{2.5cm}}\toprule
        
        \multicolumn{2}{c|}{} &Information \newline erasure \newline (incompatibility) &Distance \newline enlargement \newline (compatibility) &\textbf{Our main finding: \newline Systematic diffusion \newline (incompatibility)} &\textbf{Conclusion} \\ \cmidrule{1-6}
        
        \multicolumn{2}{l|}{\citet{muller}} &LS erases relative information in the logits & & &LS-trained teacher can hurt KD \\ \midrule
        
        \multicolumn{2}{l|}{\citet{shen2021}} &With LS, some relative information in the logits is still retained 
        &
        LS enlarges the distance between semantically similar classes 
        & 
        &
        Benefits outweigh disadvantages. LS is compatible with KD \\ \midrule
        \multirow{12}{*}{\bf Our work} 
        &
        Lower $T$ ($i.e.: T=1$) 
        &
        We agree with  \cite{shen2021} in information erasure 
        &
        We experimentally validate the inheritance of distance enlargement in the student, see Figure~\ref{fig:main}. (\citet{shen2021} has not shown this). 
        & 
        With KD of lower $T$ (i.e.: $T$=1), there is lower degree of systematic diffusion of penultimate representations towards semantically similar classes. This doesn't curtail the distance enlargement benefit. 
        &
        At lower levels of systematic diffusion in student. LS is compatible with KD \\ \cmidrule{2-6}
        &
        Increase of $T$ 
        &
        The loss of logits’ relative information cannot be recovered with an increased $T$ 
        &
        We agree with \cite{shen2021} observation, but the distance enlargement is curtailed at an increased $T$.
        &
        With KD of increased $T$, there is systematic diffusion of penultimate representations towards semantically similar classes, curtailing the distance enlargement (Sec.~\ref{sec:systematic-diffusion}).
        &
        At higher levels of systematic diffusion in student. LS and KD are not compatible. \\
        \bottomrule
        \end{tabular}
    \end{adjustbox}
    \label{table:main-findings}
\vspace{-0.5cm}
\end{table*}

%% ------------------ end ------------------

{\bf A rule of thumb for practitioners.} We suggest to use an LS-trained
teacher with a low-temperature transfer (i.e. $T$ = 1) to
achieve high performance students.

\textbf{Paper organization.} In Sec. \ref{sec:prereq}, we review LS and KD. In Sec. \ref{sec:a-closer-look-at-ls-kd}, we revisit key findings of \cite{muller} and \citet{shen2021} to emphasize the research gap. 
\emph{
Our main contribution is Sec. \ref{sec:systematic-diffusion}, where we introduce our discovered systematic diffusion, conduct qualitative, quantitative and analytical studies to verify that the diffusion is not isotopic but systematic towards semantically-similar classes, and therefore it directly curtails the benefits of using an LS-trained teacher.
}
In Sec. \ref{sec:empirical-studies}, we perform rich empirical studies to support our main finding on Systematic Diffusion. 
In Sec. \ref{sec:extended-experiments} , we conduct extended experiments using compact students and neural machine translation tasks to further support our finding.
In Sec. \ref{sec:discussion}, we provide our perspective regarding the combined use of LS and KD as empirical guidelines for practitioners, and finally conclude this study.

%%% ===============Prerequisites==================
\section{Prerequisites}
\label{sec:prereq}
%%% -------------Label Smoothing------------------
% \subsection{Label Smoothing (LS)}
% \label{prereq-ls}
\textbf{Label Smoothing (LS)} \citep{szegedy_ls}: LS was formulated as a regularization strategy to
alleviate models' over-confidence.
LS replaces the original hard target distribution with a mixture of original hard target distribution and the uniform distribution characterized by the mixture parameter $\alpha$.
Many state-of-the-art models have leveraged on LS to improve the accuracy of deep neural networks across multiple tasks including image classification \citep{he2019, aggarwal, zoph, huang}, machine translation \citep{vaswani} and speech recognition \citep{Chorowski2017, chiu, pereyra}.
Consider the formulation of LS objective with mixture
parameter $\alpha$ as follows: 
Let $p_{k}, \mathbf{w}_{k}$ represent the probability and last layer weights (including biases) corresponding to the $k$-th class. 
Let $\mathbf{x}, y_{k}, y_{k}^{LS}$ represent the penultimate layer activations, true targets and LS-targets where 
$y_{k}=1$ for the correct class and 0 for all the incorrect classes\footnote{$\mathbf{x}$ is concatenated with 1 at the end to include bias as $\mathbf{w}_{k}$ includes biases at the end.}. $\mathbf{x}^T$ is the transpose of $\mathbf{x}$.
Then for a classification network trained with LS containing $K$ classes, we minimize the cross entropy loss between LS-targets $y_{k}^{LS}$ and model predictions $p_{k}$ given by
$ L_{LS}(\mathbf{y}, \mathbf{p}) = \sum_{k=1}^{K} -y_{k}^{LS}\log(p_{k})$,
%$L_{LS}(\mathbf{y}, \mathbf{p})$. 
where 
$p_{k} = {\exp({\mathbf{x}^{T}\mathbf{w}_{k}})}/{\sum_{l=1}^{K} \exp({\mathbf{x}^{T}\mathbf{w}_{l}})}$ and $ y_{k}^{LS}=y_{k}(1-\alpha) + \frac{\alpha}{K}$.

%%% -------------Knowledge Distillation------------------
% \subsection{Knowledge Distillation (KD)}
% \label{prereq-kd}
\textbf{Knowledge distillation (KD)} \citet{hinton_kd}: KD uses a larger capacity teacher model(s)
to transfer the knowledge to a compact student model.
Recently KD methods have been widely used in visual recognition \citep{zhang_eccv20, Peng_2019_ICCV, LopSchBotVap16}, NLP \citep{hu-attention, jiao-tinybert, nakashole-flauger}
and
speech recognition \citep{KnowledgeDR_Shen, kwon, Perez_2020_WACV}. 
% and self-supervision \citep{fang2021seed}.
The success of KD methods is largely attributed to the information about incorrect classes encoded in the output distribution produced by the teacher model(s) \citep{hinton_kd}.
% \textbf{Mathematical formulation.} 
Consider KD for a classification objective. Let $T$ indicate the temperature factor that controls the importance of each soft target. Given the $k$-th class logit $\mathbf{x}^{T}\mathbf{w}_{k}$, let the temperature scaled probability be $p_{k}(T)$. For KD training, let the loss be $L_{KD}$. For $L_{KD}$, we replace the cross entropy loss $H(\mathbf{y}, \mathbf{p})$ with a weighted sum (parametrized by $\beta$) of $H(\mathbf{y}, \mathbf{p})$ and $H(\mathbf{p}^{t}(T), \mathbf{p}(T))$ where $\mathbf{p}^{t}(T), \mathbf{p}(T)$ correspond to the temperature-scaled teacher and student output probabilities. That is,
$p_{k}(T) = {\exp(\frac{{\mathbf{x}^{T}\mathbf{w}_{k}}}{T})}/{\sum_{l=1}^{K} \exp(\frac{\mathbf{x}^{T}\mathbf{w}_{l}}{T})}$
and 
$L_{KD}=(1-\beta) H(\mathbf{y}, \mathbf{p}) + \beta T^2 H(\mathbf{p}^{t}(T), \mathbf{p}(T))$. 
Following \citet{hinton_kd} $T^2$ scaling is used for the soft-target optimization as $T$ will scale the gradients approximately by a factor of $1/T^2$.
Following \citet{muller, shen2021}, we set $\beta=1$ for this study since we primarily aim to isolate and study the effects of KD. $\beta=1$ achieves good performance \citep{shen2021}.

%%% =================End of Prerequisites===============

%%% ==================Main Figure of Paper =============
\begin{figure*}[h]
\begin{adjustbox}{width=0.9\textwidth,center}
\begin{tabular}{c}
    \includegraphics[width=0.7\textwidth]{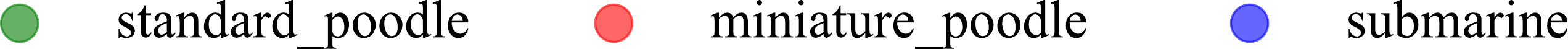} \\  
      \includegraphics[width=\textwidth]{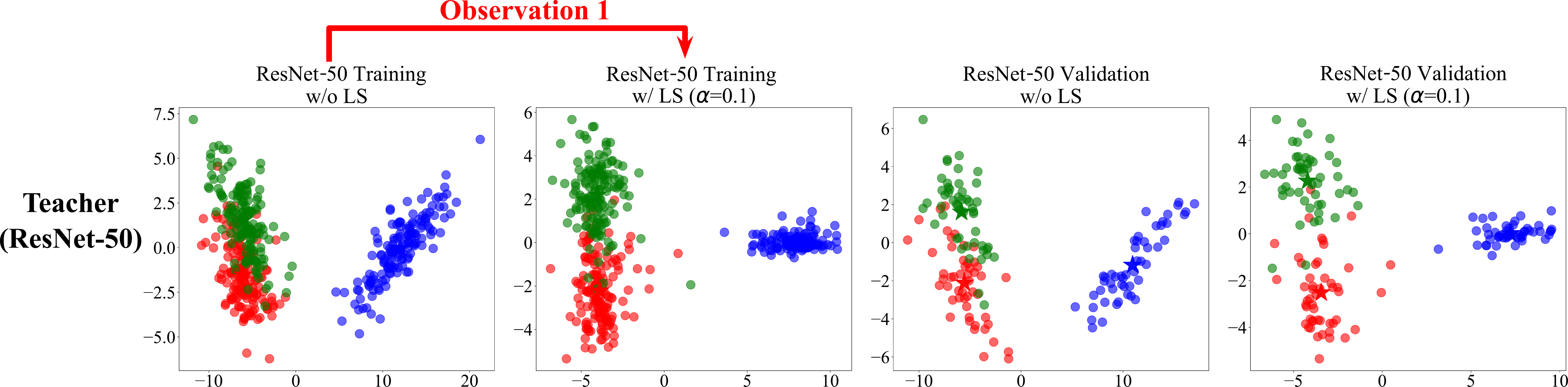} \\   
      \includegraphics[width=\textwidth]{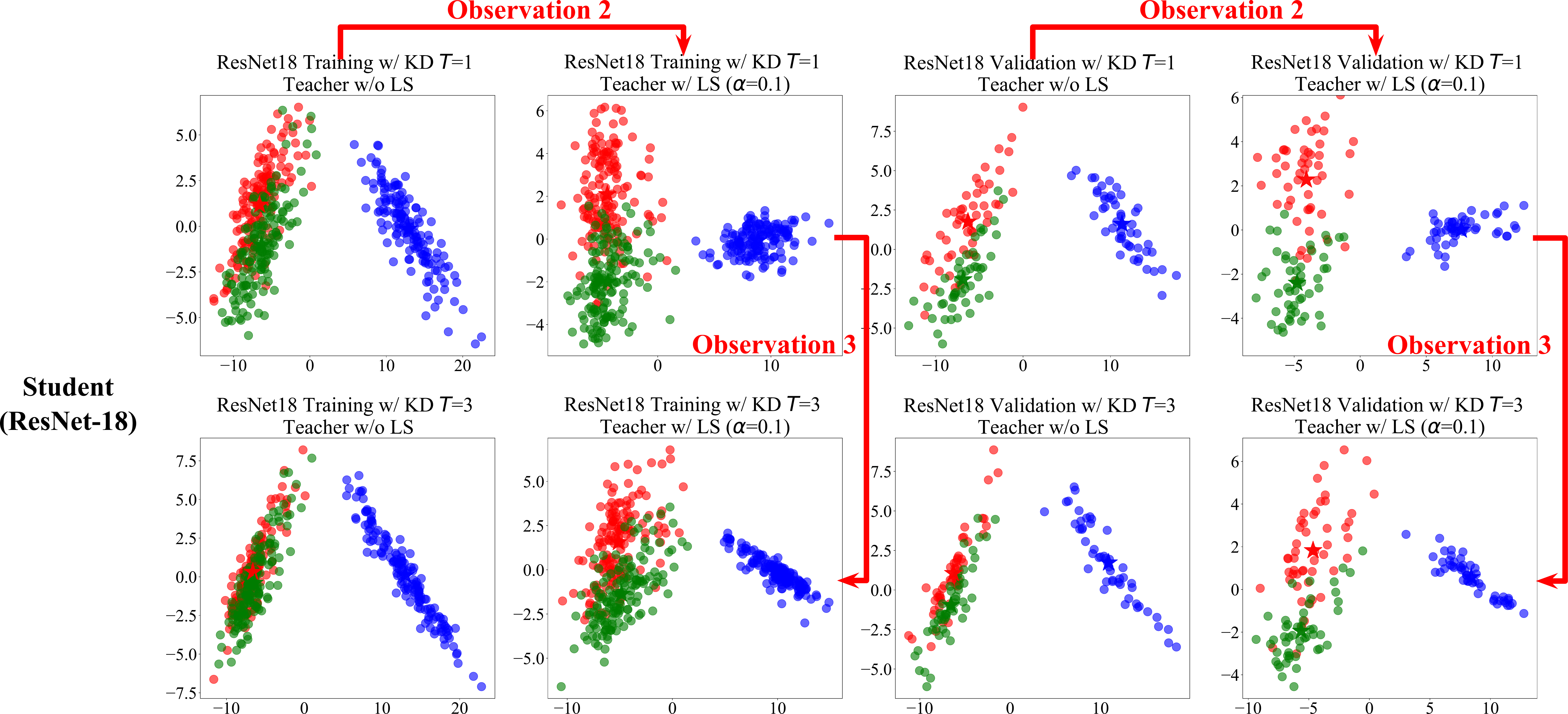} \\
\end{tabular}
\end{adjustbox}
\vspace{-0.35cm}
\caption{Visualization of the penultimate layer representations
(\texttt{Teacher = ResNet-50, Student = ResNet-18, Dataset = ImageNet}).
We follow 
the same setup and procedure used in \citet{muller}
and 
\citet{shen2021}. We also follow their three-class analysis: two semantically similar classes
({\tt \color{red} miniature\_poodle}, {\tt \color{ForestGreen} standard\_poodle})
and one semantically different class ({\tt \color{blue} submarine}).  Additional visualization can be found in the Supplementary.
{\color{red} \bf Observation 1:}
The use of LS on the teacher leads to   
tighter clusters and erasure of logits' information as claimed by \citet{muller}. In addition, increase in central distance between semantically similar classes ({\tt \color{red} miniature\_poodle},
{\tt \color{ForestGreen} standard\_poodle}) as claimed by \citet{shen2021} can be observed.
{\color{red} \bf Observation 2:}
We further visualize the student's representations. 
Increase in central distance between semantically similar classes can also be observed. This confirms the transfer of this benefit from the teacher to the student. Note that in  
\citet{muller} and \citet{shen2021}, student's representations have not been visualized.
{\color{red} \bf Observation 3 (Our main discovery):} KD of an increased $T$ causes systematic diffusion of representations between semantically similar classes ({\tt \color{red} miniature\_poodle},
{\tt \color{ForestGreen} standard\_poodle}). This curtails the increment of central distance 
between semantically similar classes
due to the use of LS-trained teacher. We notice similar observations in other datasets and networks, see Figures \ref{sup-fig:r50-imagenet}, \ref{sup-fig:r18-cub}, \ref{sup-fig:r50-cub}, \ref{sup-fig:efb0-imagenet} and \ref{sup-fig:convnext-cub}.
We also include image samples for these 3 classes in Supplementary Figure \ref{sup-fig:samples_imagenet}.
Best viewed in color.}
\label{fig:main}
\vspace{-0.5cm}
\end{figure*}

%%% ==================End of Main Figure =============

%%% ===============A Closer Look at LS and KD compatibility==================
\section{A Closer Look at LS and KD compatibility}
\label{sec:a-closer-look-at-ls-kd}
In this section, we review the contradictory findings of \citet{muller} and \citet{shen2021} from the perspective of information erasure in LS-trained teacher. This discussion is a necessary preamble
to understand our main finding, Systematic Diffusion in the student in Sec. \ref{sec:systematic-diffusion}.

% Information erasure
\textbf{Information erasure in LS-trained teacher.} 
LS objective optimizes the probability of the correct class to be equal to $1-\alpha$ + $\alpha/K$, and incorrect classes to be $\alpha/K$. 
This directly encourages the differences between logits of the correct class and incorrect classes to be a constant \citep{muller} determined by $\alpha$. 
Following \citet{muller}, the logit $\mathbf{x}^{T}\mathbf{w}_{k}$ can be approximately measured using the squared Euclidean distance between penultimate layer's activations and the template corresponding to class $k$.
That is, $\mathbf{x}^{T}\mathbf{w}_{k}$ can be approximately measured by $\Vert{\mathbf{x}-\mathbf{w}_{k}}\Vert^2$. 
This allows to establish 2 important geometric properties of LS \citep{muller}: With LS, \textit{penultimate layer activations 1) are encouraged to be close to the template of the correct class} (large logit value for the correct class, therefore small distance between the activations and the correct class template), and 2) \textit{are encouraged to be equidistant to the templates of the incorrect classes} (equal logit values for all the incorrect classes).
This results in penultimate layer activations to tightly cluster around the correct class template compared to the model trained with standard cross entropy objective. 
We demonstrate this clearly in Figure \ref{fig:main} {\color{red} \bf Observation 1}. With LS applied on the ResNet-50 model, we observe that the penultimate layer representations become much tighter.
As a result, substantial information regarding the resemblances of these instances to those of other different classes is lost. This is referred to as the information erasure in LS-trained teacher \citep{muller}.

% Why incompatible
% \textbf{Opposing claims}
\textbf{\citet{muller} finding: Information erasure in LS-trained teacher cause LS and KD to be Incompatible}: \citet{muller} are the first to investigate this compatibility, and they argue that the
information erasure effect 
due to LS (shown in Figure \ref{fig:main} {\color{red} \bf Observation 1}) can impair KD.
Given the prominent successes in KD methods being largely attributed to dark knowledge/ inter-class information emerging from the trained-teacher \citep{hinton_kd, tang2021}, 
the argument by \citet{muller} that LS and KD are incompatible due to information loss in the logits is generally convincing and widely accepted \citep{khosla, Arani_2021_WACV, tang2021, MghabbarR20, shen2021unmix}.
This is also supported by empirical evidence.

% Why incompatible
\textbf{\citet{shen2021} finding: Information erasure in LS-trained teacher provides distance enlargement benefits between semantically similar classes, resulting in LS and KD to be Compatible}: Recently an interesting finding by \citet{shen2021} argue that LS and KD are compatible. 
Though they agree that information erasure generally happens with LS, their argument focuses more on the effect of LS on semantically similar classes. 
They argue that information erasure in LS-trained teacher can promote enlargement of central distance of clusters between semantically similar classes. This allows the student network to easily learn to classify semantically similar classes which are generally difficult to classify in conventional training procedures.
We show this increased separation between semantically similar classes with LS in Figure \ref{fig:main} {\color{red} \bf Observation 1}. 
It can be observed  that the central distance between the clusters of {\tt \color{ForestGreen} standard\_poodle} and {\tt \color{red} miniature\_poodle} increases with using LS on the ResNet-50 teacher.
In our work,
we further extend to show that this property is inherited by the ResNet-18 student as well in {\color{red} \bf Observation 2}.
We remark that this inheritance is not shown by \citet{shen2021}. 
This finding by \citet{shen2021} is supported by experiments and quantitative results.
Though they claim that the benefit derived from larger separation between semantically similar classes outweigh the drawbacks due to information erasure, thereby making LS and KD compatible, their investigation does not address the contradictory findings and results reported by \citet{muller}.

\textbf{Research Gap: } Studied in isolation, both these contradictory arguments are convincing and well supported empirically. This has caused serious perplexity among the research community regarding the combined use of LS and KD.

%%% =============== End of A Closer Look at LS and KD compatibility==================

%%% ===============Systematic Diffusion in Student==================
\section{Systematic Diffusion in Student }
\label{sec:systematic-diffusion}
Through profound investigation, we discover an intriguing phenomenon occurring in the student called \textit{systematic} diffusion when distilling from an LS-trained teacher at higher $T$.
Particularly, this \textit{diffusion} maneuvers the penultimate layer representations learnt by the student of a given class in a \textit{systematic} way that targets in the direction of semantically similar classes.
This systematic diffusion is critical as it directly curtails the distance enlargement benefits between semantically similar classes when distilling from an LS-trained teacher.

\textbf{Penultimate layer visualization as evidence of systematic diffusion.}
We follow \citet{muller}, and use their visualization method based on linear projections of the penultimate layer representations.
See Figure \ref{fig:main} for visualization (We discuss Figure \ref{fig:main} deeply in Sec. \ref{sec:empirical-studies}).
Particularly, our discovery on systematic diffusion affects the distance between semantically similar classes in the student when distilled from an LS-trained teacher at higher $T$. 
This systematic diffusion can be clearly observed by visualizing the penultimate layer representations of the student.
We include the visualization algorithm and Numpy-style code in Supplementary \ref{sup-sec:visualization-algorithm}.

Given that the increased cluster center separation between semantically similar classes being the reason for the compatibility claim between LS and KD \citep{shen2021}, we discover that this cluster center separation is affected by the degree of systematic diffusion in the student.
Importantly, systematic diffusion is instrumental in explaining and resolving the contradictory findings of \citet{muller} and \citet{shen2021}, thereby establishing a foundational understanding on the compatibility between LS and KD.

\textbf{Formulation of Diffusion index ($\eta$) to measure systematic diffusion.}
To comprehensively support our discovery, we formulate a novel metric called diffusion index ($\eta$) to quantitatively measure this systematic diffusion.
% Given that this diffusion is systematic and targeted towards semantically similar classes, any quantitative metric should leverage on the use of semantics.
Given that the interpretation of `semantics' is rather subjective, we carefully construct this metric to support our discovery. 
The basic idea of this metric is to quantify the \emph{distance change} between clusters in the student network when distilled from an LS-trained teacher at higher $T$.  
\emph{
Critically, the design of the metric is to verify that the diffusion is systematic: i.e. at higher $T$, inter-cluster distance decreases for semantic similar classes and increases (relatively) for the remaining classes. As explained in the Introduction, this systematic behaviour is critical in our study.
}
There are important considerations in formulating this metric discussed below.

\vspace{-0.45cm}
\begin{itemize}[leftmargin=0.5cm]
    \setlength{\itemsep}{1pt}
    \setlength{\parsep}{1pt}
    \setlength{\parskip}{1pt}
    \item A target class $\pi$ can be characterized by the centroid of the penultimate layer representations of samples belonging to $\pi$. Let the centroid of class $\pi$ be $c_{\pi}$.
    
    \item Consider the sets $S_1, S_2$ where $S_1$ contains $|S_1|$ semantically similar classes to $\pi$ and $S_2$ contains $|S_2|$ semantically dissimilar classes to $\pi$. $|S|$ indicates the number classes in the set $S$. For easier understanding, consider 2 classes $p, q$ where $p \in S_1, q \in S_2$.
    
    \item The proximity of $c_{\pi}$ to $c_p$ can approximately measure the semantic similarity between class $\pi$ and $p$. Though this proximity can be directly measured by Euclidean distance between centroids, it requires some careful thought on normalization. The reason is as follows: What we are interested is how close is centroid of class $\pi$ to class $p$ compared to class $q$. In other words, we are interested in the {\em relative} distance between centroids of classes $(\pi, p)$ and $(\pi, q)$. Hence to measure this relative distance we normalize the distance by the sum of pairwise distance from $c_\pi$ to centroids of all other classes in $S$.
    
    \item Do note that the location of the centroids will change with temperature. In fact, we are interested in the change of centroids with increased $T$ to measure this systematic diffusion. We formulate the following diffusion index $\eta$ to measure the average percentage change in distances between semantically similar classes and semantically dissimilar classes with respect to a target class.
    
\end{itemize}
\vspace{-0.45cm}

Given a class $\pi$ and its centroid $\mathbf{c}_{\pi}$. Let the centroid of a class $k$ be represented by $\mathbf{c}_k$, $k \in {S_1, S_2}$. Let the temperature be $T$. We quantify the relative distance between classes $\pi$ and $k$: 

$d(\mathbf{c}_{\pi}(T), \mathbf{c}_k(T)) = \frac{ \Vert \mathbf{c}_{\pi}(T) - \mathbf{c}_k(T) \Vert^2 }{R}$,
where $R =\sum_{p \in S_1} \Vert \mathbf{c}_{\pi}(T) - \mathbf{c}_{p}(T) \Vert^2 + \sum_{q \in S_2} \Vert \mathbf{c}_{\pi}(T) - \mathbf{c}_{q}(T) \Vert^2$ (normalization constant). The diffusion index $\eta$ measures the average percentage change in distance between a target class $\pi$ and classes in the set $S$ \textit{when temperature is changed from $T_1$ to $T_2$} defined as follows:
\begin{equation}
    \eta(T_1, T_2; \pi, S) = \frac{1}{|S|} \sum_{k \in S}\frac{  \delta }{d(\mathbf{c}_{\pi}(T_1), \mathbf{c}_{k}(T_1))},
\label{eqn}
\end{equation}
where $\delta=d(\mathbf{c}_{\pi}(T_2), \mathbf{c}_{k}(T_2)) - d(\mathbf{c}_{\pi}(T_1), \mathbf{c}_{k}(T_1))$. Substituting $S_1$, $S_2$ into $S$ of Eq. \ref{eqn}, we have: i) $\eta(T_1, T_2; \pi, S_1)$ measures the change in relative distance between class $\pi$ and its semantically \textit{similar} class in $S_1$. ii) $\eta(T_1, T_2; \pi, S_2)$ measures the change in relative distance between class $\pi$ and its semantically \textit{dissimilar} class in $S_2$. 
We discuss empirical results for $\eta$ in Sec. \ref{sec:empirical-studies}

To give more intuition on $\eta$, consider the 3 class example (Figure \ref{fig:main}): {\tt \color{red} miniature\_poodle} (as $\pi$ class), {\tt \color{ForestGreen} standard\_poodle} (as $p \in S_1$ and $|S_1|=1$), {\tt \color{blue} submarine} (as $q \in S_2$ and $|S_2|=1$). As $T$ increases from $T_1=1$ to $T_2=3$, the relative distance between {\tt \color{red} miniature\_poodle} and {\tt \color{ForestGreen} standard\_poodle} reduces due to diffusion (Figure \ref{fig:main}), therefore $d(\mathbf{c}_{\pi}(T_2), \mathbf{c}_{p}(T_2)) < d(\mathbf{c}_{\pi}(T_1), \mathbf{c}_{p}(T_1))$. From Eq. \ref{eqn}, it is clear that the numerator will be negative. We normalize by the reference distance to calculate the percentage change. As a result, the average percentage change over $S_1$ will be negative, indicating diffusion towards semantically similar classes. 
Similarly when measured over $S_2$, the average percentage change between {\tt \color{red} miniature\_poodle} and {\tt \color{blue} submarine} will be positive (because $d(\mathbf{c}_{\pi}(T_2), \mathbf{c}_{q}(T_2)) > d(\mathbf{c}_{\pi}(T_1), \mathbf{c}_{q}(T_1))$ as observed in  Figure \ref{fig:main}) indicating diffusion away from $\pi$.
% the target class.

\begin{figure*}
    \centering
    \includegraphics[width=0.98\textwidth]{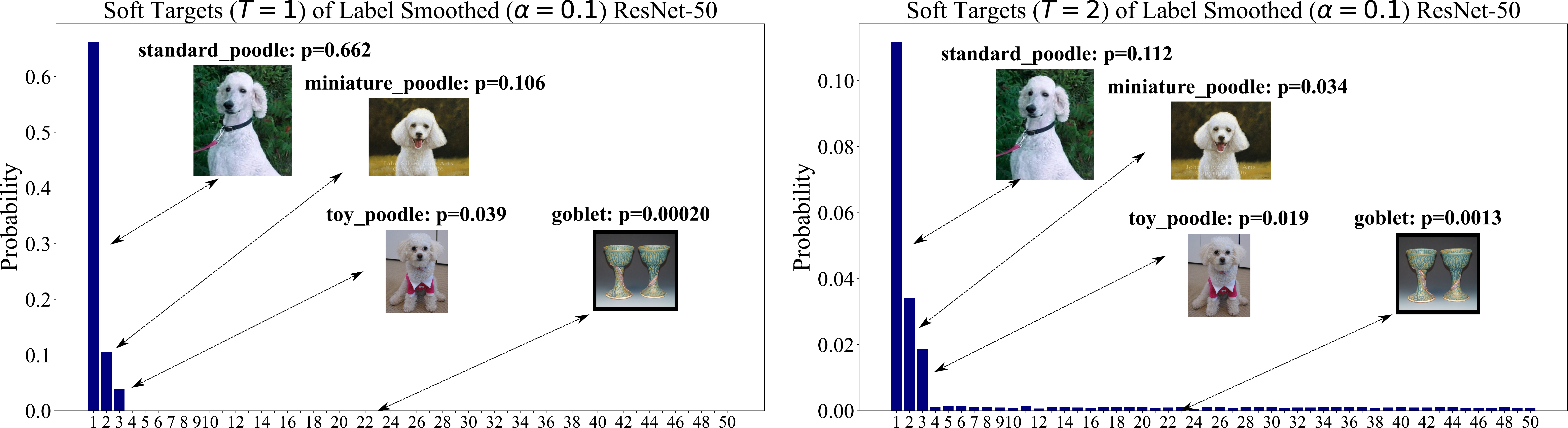}
    \vspace{-0.35cm}
    \caption{\footnotesize Soft output of the LS-trained ResNet-50 teacher $(\alpha =0.1)$ same as the one in Figure~\ref{fig:main}.
    Left: soft output at $T=1$; Right: soft output at $T=2$.
    The figures show the average of the soft outputs for 
    1300 training {\tt standard\_poodle} samples.
    Index 1 is the soft output for the {\tt standard\_poodle} class, i.e. 
    ${p}^t_{k^*}(T)$. Index 2 and 3 are the soft outputs for the semantically similar classes 
    {\tt miniature\_poodle} and {\tt toy\_poodle} respectively, i.e. ${p}^t_{ml}(T)$.
    The rest are soft outputs of randomly-chosen semantically dissimilar classes, i.e. ${p}^t_{ms}(T)$.
    Note that an increase of $T$ brings ${p}^t_{ml}(T)$ closer to ${p}^t_{k^*}(T)$. Therefore, soft targets at an increased $T$ encourage student to learn penultimate representations closer to semantically similar class $ml$, which are {\tt miniature\_poodle} and {\tt toy\_poodle} in this case. Therefore, in Figure~\ref{fig:main} {\color{red}Observation 3}, {\tt standard\_poodle} activations has more overlapping with {\tt miniature\_poodle} when KD of $T=2$ is used.
    Also, ${p}^t_{ms}(T)$ remains negligible after $T$ scaling, as shown in the figure.
    Furthermore, the  figure of $T=1$ (Left) suggests that even with LS 
    probabilities of incorrect classes
    $\{{p}^t_{m}\}$ 
    are not all the same, and information erasure is not prefect in practice.
    Therefore, the diffusion of penultimate representations is not isotopic.}
    \label{figure_2_logit}
    \vspace{-0.4cm}
\end{figure*}

\textbf{Why is this diffusion systematic and not isotopic?}
We revisit discussion from \citet{hinton_kd} to motivate the intuition behind this \emph{systematic} diffusion. \citet{hinton_kd} introduce $T$ to scale the logits at the final softmax in order to produce soft targets that are more suitable for transfer. As argued by \citet{hinton_kd} on MNIST classification, a sample of `2' may be assigned a probability of $10^{-6}$ of being a `3' and $10^{-9}$ of being a 7. The resemblance between `2' and `3' is valuable information, but a probability of $10^{-6}$ has negligible influence on the loss when distilling to student. \citet{hinton_kd} introduce a temperature $T$ to emphasize the probabilities of such incorrect classes: during KD, their $T$-scaled counterparts have more noticeable effects on the student. On the other hand, the effect of $T$ scaling on the probability of $10^{-9}$ is negligible; consequently, the $T$-scaled counterparts of such probabilities remain to have unnoticeable effects on the student.

In particular, for a given sample of ground-truth class $k^*$, we let ${p}^t_{k^*}$ represent the  probability of the correct class output by the teacher, ${p}^t_{m}$ represent the probability of one of the $K-1$ incorrect classes.
Among these $K-1$ ${p}^t_{m}$, one or a few could be significantly larger than the other; we refer such probability as ${p}^t_{ml}$ (i.e.: probability of being a 3 in the above example). 
In particular, the class $ml$ is usually a semantically similar class of class $k^*$, therefore 
${p}^t_{ml}$ is not negligible for a class $k^*$ sample
(See Figure~\ref{figure_2_logit}).
For the rest of 
${p}^t_{m}$ which are almost zero (noise level), we refer them as ${p}^t_{ms}$ (e.g., probability of being a 7 in the above example). Therefore, $\{{p}^t_{m}\} = \{{p}^t_{ml}\} \cup \{{p}^t_{ms}\}$. 
Usually, we have
%(i) ${p}^t_{k^*} > {p}^t_{ml}$
${p}^t_{ml} \gg {p}^t_{ms}$ and
${p}^t_{ms} \approx 0$.
%(iv) ${p}^t_{ms}$ are the majorities.
We remark that 
$\{{p}^t_{m}\}$ are not all the same and can be observed even for an LS-trained teacher. It is because logits' information is not completely erased  (see Figure~\ref{figure_2_logit}).

When KD of an increased $T$ is used, the soft output of the teacher is scaled and becomes $\mathbf{p}^t(T)$. In particular, the effect of $T$ scaling is
to bring ${p}^t_{ml}$ closer to  
${p}^t_{k^*}$, i.e., 
${p}^t_{ml}(T)$ is closer to 
${p}^t_{k^*}(T)$ relatively.
Consequently, 
with soft target $\mathbf{p}^t(T)$,
student is 
encouraged to produce a penultimate representation
of a class $k^*$ sample
that is  closer 
to the incorrect class $ml$. This results  in systematic diffusion
of representations of class $k^*$
towards the incorrect class $ml$.
This can be observed in Figure~\ref{fig:main} {\color{red} Observation 3} for {\tt \color{ForestGreen} standard\_poodle} activations (here class $ml$ being {\tt \color{red} miniature\_poodle}), and similarly for {\tt \color{red} miniature\_poodle} activations.
On the other hand, because ${p}^t_{ms}$ is negligibly small, 
even with $T$ scaling
${p}^t_{ms}(T)$ remains negligible and has unnoticeable effect for student's penultimate representation. Therefore, the diffusion due to an increased $T$ is not isotopic but towards semantically similar classes (class $ml$).  
We provide more detailed discussion on systematic diffusion in Supplementary \ref{sup-sec:additional-systematic-diffusion}.

We remark that this systematic diffusion can sometimes be observed when using a  teacher without LS, see 
Figure~\ref{fig:main}, row 2 subplot 1 and row 3 subplot 1.
For a teacher without LS (i.e. no information erasure), this 
systematic diffusion
could in fact be advantageous in some cases, as it improves generalization of the student network using the rich logits' information about instance resemblances.
{\em However, we focus on our thesis statement: compatibility between LS and KD. In our case, 
systematic diffusion in student due to KD at an increased $T$ curtails the distance enlargement (between semantically similar classes) benefits of using an LS-trained teacher, rendering KD ineffective.}

%%% ===============End of Systematic Diffusion in Student==================

%%% ===============Empirical Studies=================
\section{Empirical Studies}
\label{sec:empirical-studies}

In this section, we conduct large-scale 
image classification (standard, fine-grained)
LS-KD experiments.
We remark that LS and KD are compatible when with all the other factors fixed (including $T$), student distilled from an LS-trained teacher \textit{outperforms} the student distilled from a teacher trained without LS. 
We use ResNet-50 teacher and ResNet-18, ResNet-50 students
using ImageNet-1K and CUB200-2011 datasets following similar procedure as \citet{shen2021}. 
Results are shown in Table \ref{table:results-imagenet-cub}.

%%------- ImageNet-1K Results------------------
\begin{table*}[h]
    \caption{
    % Knowledge distillation 
    KD
    results from ResNet-50 Teacher to ResNet-18, ResNet-50 students
    for (A) standard image classification using ImageNet-1K and (B) fine-grained image classification using CUB200-2011 benchmarks 
    following similar procedure as \citet{shen2021}.
    We show the top1/ top5 test accuracies. 
    Configurations where LS and KD are compatible are in \textbf{bold}.
    As one can clearly observe, \emph{with LS-trained teacher, there is a consistent degrade in student performance as $T$ increases. This can be observed in all our 34 experiments.}
    These results comprehensively support our claim: \emph{in the presence of an LS-trained teacher, KD at higher temperatures is rendered ineffective.}
    On the other hand, we observe that higher $T$ can improve the performance of ResNet-18 student when using a teacher trained without LS in fine-grained classification using CUB200-2011 (B). 
    i.e.: We observe improvement of ResNet-18 student from $T=1$ to $T=2$, $T=3$ when distilling from teacher trained without LS in (B). 
    % and compact student network distillation experiments (See Supplementary Tables \ref{sup-table:r18-cub} and \ref{sup-table:mobilenetv2-cub}).
    We particularly emphasize that our findings are exclusive to LS and KD: That is in the presence of an LS-trained teacher, higher $T$ renders ineffective KD due to systematic diffusion.
    All these results are averaged over 3 independent runs. Standard deviations are reported in Supplementary Tables \ref{table:sup-std-imagenet}, \ref{table:sup-std-cub} respectively.
    }
    \begin{tabular}{cc}
    \begin{minipage}{\columnwidth}
    \begin{adjustbox}{width=0.95\columnwidth,center}
        \begin{tabular}{l|c|c|c}
        \multicolumn{4}{c}{A. ImageNet-1K : ResNet-50 to ResNet-18, ResNet-50 KD}\\
        \toprule
        &\backslashbox{$T$}{$\alpha$} &$\alpha$ = 0.0 &$\alpha$ = 0.1 \\ \midrule
        Teacher : ResNet-50 &- &76.130 / 92.862 &76.196 / 93.078 \\ \midrule
        \multirow{4}{*}{Student : ResNet-18} &$T$ = 1 &71.547 / 90.297 &\textbf{71.616 / 90.233} \\ \cmidrule{2-4}
        &$T$ = 2 &71.349 / 90.359 &68.428 / 89.139 \\ \cmidrule{2-4}
        &$T$ = 3 &69.570 / 89.657 &66.570 / 88.631 \\ \cmidrule{2-4}
        &$T$ = 64 &66.230 / 88.730 &65.472 / 89.564 \\
        \midrule
        \multirow{4}{*}{Student : ResNet-50} &$T$ = 1 &76.502 / 93.059 &\textbf{77.035 / 93.327} \\ \cmidrule{2-4}
        &$T$ = 2 &76.198 / 92.987 &76.101 / 93.115 \\ \cmidrule{2-4}
        &$T$ = 3 &75.388 / 92.676 &\textbf{75.821 / 93.065} \\ \cmidrule{2-4}
        &$T$ = 64 &74.291 / 92.399 &\textbf{74.627 / 92.639} \\
        \bottomrule
        \end{tabular}
    \end{adjustbox}
    \label{table:results-imagenet-resnet18}
    \end{minipage}
    
    &
    
    \begin{minipage}{\columnwidth}
    \begin{adjustbox}{width=0.95\columnwidth,center}
        \begin{tabular}{l|c|c|c}
        \multicolumn{4}{c}{B. CUB200-2011 : ResNet-50 to ResNet-18, ResNet-50 KD}\\ \toprule
        &\backslashbox{$T$}{$\alpha$} &$\alpha$ = 0.0 &$\alpha$ = 0.1 \\ \toprule
        Teacher : ResNet-50 &- &81.584 / 95.927 &82.068 / 96.168 \\ \midrule
       \multirow{4}{*}{Student : ResNet-18} &$T$ = 1 &80.169 / 95.392  &\textbf{80.946 / 95.312} \\ \cmidrule{2-4}
        &$T$ = 2 &80.808 / 95.593 &80.428 / 95.518 \\ \cmidrule{2-4}
        &$T$ = 3 &80.785 / 95.674 &78.196 / 95.213 \\ \cmidrule{2-4}
        &$T$ = 64 &73.611 / 94.529 &67.161 / 93.062 \\
        \midrule
        \multirow{4}{*}{Student : ResNet-50} &$T$ = 1 &82.902 / 96.358 &\textbf{83.742 / 96.778} \\ \cmidrule{2-4}
        &$T$ = 2 &82.534 / 96.427 &\textbf{83.379 / 96.537} \\ \cmidrule{2-4}
        &$T$ = 3 &82.091 / 96.243 &\textbf{82.142 / 96.427} \\ \cmidrule{2-4}
        &$T$ = 64 &79.784 / 95.927 &77.206 / 95.812 \\
        \bottomrule
        \end{tabular}
        
    \end{adjustbox}
    \label{table:results-cub-resnet18}
    \end{minipage}
    \vspace{0.3cm}
    \\
    
    \end{tabular}
    \label{table:results-imagenet-cub}
\vspace{-0.3cm}
\end{table*}
%%------- End of ImageNet-1K Results------------------

\textbf{Penultimate layer visualization analysis.}
We show this systematic diffusion in ResNet-18 student using Figure \ref{fig:main} {\color{red} \bf Observation 3}. 
We focus on the two semantically similar classes: 
{\tt \color{red} miniature\_poodle},
{\tt \color{ForestGreen} standard\_poodle}.
Given the same LS-trained ResNet-50 teacher and using the exact distillation process, we observe that at increased temperatures ($T=1$ to $T=3$), the above semantically similar classes start to diffuse. We also observe that  class {\tt \color{blue} submarine} diffuses towards another class which is semantically similar to {\tt \color{blue} submarine} (not shown in the figure).
Because of this systematic diffusion, the central cluster distances between {\tt \color{red} miniature\_poodle} and 
{\tt \color{ForestGreen} standard\_poodle} reduces with increased $T$ in the presence of LS-trained teacher. Consequently, this systematic diffusion results in detrimental performance in the student causing an accuracy drop of 5.05\% as shown in Table \ref{table:results-imagenet-resnet18} A. 
Supporting visualization showing systematic diffusion in ResNet-50 student shown in Figure \ref{sup-fig:r50-cub} corresponding to the 1.21\% drop as shown in Table \ref{table:results-imagenet-cub}.
CUB200-2011 visualization for ResNet-18 and ResNet-50 students shown in Figures \ref{sup-fig:r18-cub}, \ref{sup-fig:r50-cub}  respectively.

\textbf{Analysis using diffusion index ($\eta$).}
We quantitatively illustrate systematic diffusion in the ResNet-18, ResNet-50 students using $\eta$ for 10 target classes in Table \ref{table:eta}. 
We clearly observe that $\eta(T_1=1, T_2=3; \pi, S_1) < 0$ and $\eta(T_1=1, T_2=3; \pi, S_2) > 0$ for all these 10 target classes, thereby quantitatively showing that the penultimate layer representations are diffused towards semantically similar classes when distilled from an LS-trained teacher at a larger temperature. 
This systematic diffusion results in detrimental performance of the student resulting in an accuracy drop of 5.05\%, 1.21\% for ResNet-18 and ResNet-50 students respectively as shown in Table \ref{table:results-imagenet-resnet18} A. 
We also include a rich study on selecting $S_1$ and $S_2$ in Supplementary \ref{sup-sec:semantic-classes}.
% We show similar analysis for ResNet-50 student in Supplementary \ref{sup-sec:image-classification}. 

\begin{table*}[!h]
    \vspace{-0.3cm}
    \caption{\footnotesize
    $\eta$ analysis for ResNet-18 (top), ResNet-50 (bottom) students for 10 target classes in ImageNet-1K (We show in 2 sets). 
    % We use ImageNet hierarchy derived from WordNet \citep{Fellbaum1998} to select 4 semantically similar classes and 20 semantically dissimilar classes (random) to compute the diffusion index $\eta$. $|S_1|=4$ and $|S_2|=20$ for each target class.
    We use standard, pre-defined ImageNet-1K knowledge graph derived from WordNet \citep{Fellbaum1998} as a prior to select 4 semantically similar classes and 20 semantically dissimilar classes (random) to compute the diffusion index $\eta$. $|S_1|=4$ and $|S_2|=20$ for each target class. 
    % Do note that we also verify that the classes in $S_1$ are the closest 1\% of the classes to the target class and classes in $S_2$ are among the distant 90\% of the classes to the target class using feature space distance measurements.
    %
    We demonstrate that when increasing $T=1$ to $T=3$, the diffusion index $\eta$ between target class and $S_1$ reduces substantially and vice versa for $S_2$ shown for both training and validation set. 
    This quantitatively shows systematic diffusion when distilling at higher $T$ in the presence of an LS-trained teacher.
    }
    
    \begin{tabular}{cc}
    \begin{minipage}{\columnwidth}
    \begin{adjustbox}{width=0.95\columnwidth,center}
       \begin{tabular}{lcccc}
        \multicolumn{5}{c}{\large{Set 1 : ResNet-18 student}}\\
        \toprule
        Target class &$Train:S_1$ &$Train:S_2$ &$Val:S_1$ &$Val:S_2$ \\ \toprule
        Chesapeake Bay retriever &-0.392 &0.162 &-1.082 &0.269 \\ \midrule
        curly-coated retriever &-0.578 &0.179 &-2.024 &0.383 \\ \midrule
        flat-coated retriever &-1.729 &0.380 &-3.320 &0.655 \\ \midrule
        golden retriever &-0.880 &0.228 &-2.594 &0.555 \\ \midrule
        Labrador retriever &-2.758 &0.501 &-4.618 &0.840 \\
        \bottomrule
        \end{tabular}
    
    \end{adjustbox}
    \label{table:eta_set1}
    \end{minipage}
    
    &
    
    \begin{minipage}{\columnwidth}
    \begin{adjustbox}{width=0.90\columnwidth,center}
       \begin{tabular}{lcccc}
        \multicolumn{5}{c}{\large{Set 2 : ResNet-18 student}}\\
        \toprule
        Target class &$Train:S_1$ &$Train:S_2$ &$Val:S_1$ &$Val:S_2$ \\ \toprule
        thunder snake &-2.316 &0.376 &-3.584 &0.511 \\ \midrule
        ringneck snake &-0.463 &0.058 &-0.757 &0.094 \\ \midrule
        hognose snake &-1.528 &0.258 &-4.067 &0.631 \\ \midrule
        water snake &-2.028 &0.326 &-3.053 &0.478 \\ \midrule
        king snake &-2.474 &0.521 &-4.577 &0.840 \\
        \bottomrule
        \end{tabular}
    \end{adjustbox}
    \label{table:eta_set2}
    \vspace{0.1cm}
    \end{minipage}
    
    \\
    
    % ResNet-50
    %\renewcommand\thetable{A}
    \begin{minipage}{\columnwidth}
    %\caption{Set 1}
    \begin{adjustbox}{width=0.95\columnwidth,center}
      \begin{tabular}{lcccc}
        \multicolumn{5}{c}{\large{Set 1 : ResNet-50 student}}\\
        \toprule
        Target class &$Train:S_1$ &$Train:S_2$ &$Val:S_1$ &$Val:S_2$ \\ \toprule
        Chesapeake\_Bay\_retriever &-1.061 &0.180 &-1.346 &0.240 \\ \midrule
        curly-coated\_retriever &-0.764 &0.127 &-1.193 &0.207 \\ \midrule
        flat-coated\_retriever &-0.983 &0.169 &-0.331 &0.056 \\ \midrule
        golden\_retriever &-0.744 &0.159 &-0.911 &0.182 \\ \midrule
        Labrado\_retriever &-1.336 &0.236 &-1.468 &0.257 \\
        \bottomrule
        \end{tabular}
    
    \end{adjustbox}
    \label{sup-table:eta_set1}
    \end{minipage}
    
    &
    
    \begin{minipage}{0.44\textwidth}
    %\renewcommand\thetable{B}
    %\caption{
    %Set 2
    %}
    \begin{adjustbox}{width=\columnwidth,center}
      \begin{tabular}{lcccc}
        \multicolumn{5}{c}{\large{Set 2 : ResNet-50 student}}\\
        \toprule
        Target class &$Train:S_1$ &$Train:S_2$ &$Val:S_1$ &$Val:S_2$ \\ \toprule
        thunder snake &-2.565 &0.417 &-0.778 &0.105 \\ \midrule
        ringneck snake &-2.224 &0.358 &-0.726 &0.102 \\ \midrule
        hognose snake &-3.748 &0.623 &-2.173 &0.342 \\ \midrule
        water snake &-1.631 &0.258 &-0.390 &0.037 \\ \midrule
        king snake \footnotemark &-1.969 &0.339 &0.956 &-0.159 \\
        \bottomrule
        \end{tabular}
    
    \end{adjustbox}
    \label{sup-table:eta_set2}
    \end{minipage}

    \end{tabular}
    \label{table:eta}

\vspace{-0.2cm}
\end{table*}

\textbf{Resolving the contradictory claims using systematic diffusion.}
The seemingly contradictory findings of \citet{muller} and \citet{shen2021} can be resolved using our discovery on systematic diffusion as follows: 
\citet{muller} make the incompatibility claim between LS and KD due to observing students distilled from LS-trained teacher performing inferior to students distilled from teacher trained without LS \textit{at higher $T$}. 
On the contrary, \citet{shen2021} make the compatibility claim between LS and KD due to observing students distilled from LS-trained teacher performing superior to students distilled from teacher trained without LS \textit{at lower $T$ (i.e.: $T=1$)}. 
Critically, our main finding shows that \textit{in  the  presence  of  an  LS-trained  teacher,  KD  at  higher  temperatures  systematically  diffuses penultimate layer representations learnt by the student towards semantically similar classes. 
This systematic diffusion essentially curtails the distance enlargement (between semantically similar classes) benefits of distilling from an LS-trained teacher, thereby rendering KD at increased temperatures ineffective.}
More specifically, in the presence of an LS-trained teacher, the degree of systematic diffusion is low when 
distilling at lower $T$
thereby making LS and KD compatible. 
On the other hand, the degree of systematic diffusion is relatively higher when distilling at higher $T$, thereby making LS and KD incompatible. 
% We clearly show this using penultimate layer visualization and $\eta$ analysis.
Our findings are summarized in Table \ref{table:main-findings}.
Importantly, 
% we remark that 
systematic diffusion was the missing concept that is instrumental in resolving the contradictory claims of \citet{muller} and \citet{shen2021}.

\footnotetext{For king snake target class, $\eta(T_1=1, T_2=3; \pi, S_1) < 0$ for training set and not validation. We remark that training set is used during distillation.}

\section{Extended Experiments}
\label{sec:extended-experiments}

\textbf{Compact Student Distillation.}
KD is one of the most prominent methods used for neural network compression.
% (i.e.: edge intelligence, mobile applications).
Hence, we conduct KD experiments to transfer knowledge 
% from a large, powerful teacher model  
to compact student model.
We conduct fine-grained classification experiments (CUB200-2011) using ResNet-50 teacher (25.6M params) and MobileNet-V2 student (3.50M params).
The results are shown in Table \ref{table:mobilenetv2-cub}.
Our results show that in the presence of an LS-trained teacher, KD at higher temperatures is rendered ineffective due to systematic diffusion in the student.
% Penultimate layer visualizations shown in Supplementary  
We also show supporting results for EfficientNet-B0 (for ImageNet-1K classification): Table \ref{sup-table:efb0-imagenet}. Visualization : Figure \ref{sup-fig:efb0-imagenet} and $\eta$ results : Table \ref{sup-table:efb0-imagenet-eta}.
\vspace{-0.01cm}

\begin{table}[h]
    \centering
    \vspace{-0.65cm}
    \caption{
    \textbf{Compact student distillation results:}
    Top1/ Top5 Accuracy for KD experiments from  ResNet-50 Teacher to MobileNetV2 student using CUB200-2011.
    Configurations where LS and KD are compatible are in \textbf{bold}.
    These results support our claim: \emph{in the presence of an LS-trained teacher, KD at higher temperatures is rendered ineffective.}
    We also observe that higher $T$ is helpful when distilling from a teacher trained without LS in this setup (Observe improvement of student from $T=1$ to $T=2$, $T=3$ when distilling from teacher trained without LS). 
    Standard deviations reported in Table \ref{sup-table:mobilenetv2-cub-std}.
    }
        
    \begin{adjustbox}{width=0.99\columnwidth,center}
        \begin{tabular}{l|c|c|c}\toprule
        &\backslashbox{$T$}{$\alpha$} &$\alpha$ = 0.0 &$\alpha$ = 0.1 \\ \toprule
        Teacher : ResNet-50 &- &81.584 / 95.927 &82.068 / 96.168 \\ \midrule
        \multirow{4}{*}{Student : MobileNet-V2} &$T$ = 1 &81.144 / 95.677 &\textbf{81.731 / 95.754} \\ \cmidrule{2-4}
        &$T$ = 2 &81.895 / 95.858 &80.609 / 95.47 \\ \cmidrule{2-4}
        &$T$ = 3 &81.257 / 95.677 &78.961 / 95.306 \\ \cmidrule{2-4}
        &$T$ = 64 &75.441 / 94.702 &70.435 / 93.494 \\
        \bottomrule
        \end{tabular}
    \end{adjustbox}
\label{table:mobilenetv2-cub}
\end{table}

\vspace{-0.6cm}
\textbf{Neural machine translation.}
Following \citet{shen2021}, we conduct KD experiments for neural machine translation task using IWSLT dataset. We report English $\rightarrow$ German translation results in Table \ref{table:nmt-en-de}.
Our results comprehensively show that in the presence of an LS-trained teacher, KD at higher temperatures is rendered ineffective due to systematic diffusion in the student.
We also show supporting results for English $\rightarrow$ Russian translation in Table \ref{sup-table:nmt-en-ru}.
\vspace{-0.8cm}

\begin{table}[!h]
    \centering
    \vspace{-0.1cm}
    \caption{
    \textbf{Neural Machine Translation results:}
    BLEU scores for KD experiments for Transformer Teacher to Transformer student on IWSLT dataset using English $\rightarrow$ German translation task. 
    Configurations where LS and KD are compatible are in \textbf{bold}.
    These results comprehensively support our claim: \emph{in the presence of an LS-trained teacher, KD at higher temperatures is rendered ineffective.}
    Standard deviations reported in Table \ref{table:nmt-en-de-std}.
    }
        
    \begin{adjustbox}{width=0.77\columnwidth,center}
        \begin{tabular}{l|c|c|c}\toprule
        &\backslashbox{T}{$\alpha$} &$\alpha$ = 0.0 &$\alpha$ = 0.1 \\ \toprule
        Teacher : Transformer &- &26.461 &26.750 \\ \midrule
        \multirow{4}{*}{Student : Transformer} &$T$ = 1 &24.914 &\textbf{25.085} \\ \cmidrule{2-4}
        &$T$ = 2 &23.103 &\textbf{23.421} \\ \cmidrule{2-4}
        &$T$ = 3 &21.999 &\textbf{22.076} \\ \cmidrule{2-4}
        &$T$ = 64 &6.564 &6.461 \\
        \bottomrule
        \end{tabular}

    \end{adjustbox}
    \vspace{-0.8cm}
\label{table:nmt-en-de}
\end{table}

%% ----------------Discussion and Conclusion---------------------
\section{Discussion and Conclusion}
\label{sec:discussion}

\textbf{Discussion.} 
While increased $T$ is believed to be a helpful empirical trick (Also observed in many of our experiments when distilling from a teacher trained without LS) to produce better soft-targets for KD, we convincingly show that in the presence of LS-trained teacher, an increased $T$ causes systematic diffusion in the student.
% of penultimate layer representations towards semantically similar classes in the student. 
This systematic diffusion directly curtails the distance enlargement (between semantically similar classes) benefits of an LS-trained teacher, thereby rendering KD ineffective at increased $T$. 
% \textit{As a rule of thumb, we suggest to use lower $T$ (i.e.: $T=1$) for KD in the presence of an LS-trained teacher to avoid systematic diffusion to effectively take advantage of LS-trained teachers.}
\textit{For practitioners, as a rule of thumb, we suggest to use an LS-trained teacher with a low-temperature transfer (i.e. $T=1$) to render high performance students.}
We also remark that our finding on systematic diffusion substantially reduces the search space for the intractable parameter $T$ when using an LS-trained teacher.
Our findings are summarized in Table \ref{table:main-findings}.
With increasing use of KD, we hope that our findings can benefit various applications including 
neural architecture search \citep{li2020block, yu2020bignas, alphanet-wang21i}, self-supervised learning \citep{fang2021seed, abbasi2020compress}, compact deepfake / anomaly detection \citep{dzanic2020fourier, Chandrasegaran_2021_CVPR, lim2018doping, tran2021data} and GAN compression \citep{li2020gan, fu2020autogan, ssl_gan_compress}.

\textbf{Conclusion.} 
Focusing on the compatibility between LS and KD, we have conducted an empirical study to investigate the seemingly contradictory findings of \citet{muller} and \citet{shen2021}. Through comprehensive scrutiny of these works, we discover an intriguing phenomenon called \textit{systematic} diffusion: 
That is \textit{in the presence of an LS-trained teacher, KD at higher temperatures systematically diffuses penultimate layer representations learnt by the student towards semantically similar classes.  
This systematic diffusion essentially curtails the benefits of distilling from an LS-trained teacher, thereby rendering KD at increased temperatures ineffective.}
We showed this systematic diffusion both qualitatively and quantitatively using extensive analysis. We also supported our findings with large scale experiments including image classification (standard, fine-grained), neural machine translation and compact student distillation tasks. 
\textit{Critically, our discovery on systematic diffusion was the missing concept that is instrumental in resolving the contradictory findings of \citet{muller} and \citet{shen2021}}, thereby establishing a foundational understanding on the compatibility between LS and KD.
% Code /reproducibility details are included in the submission.

\textbf{Acknowledgements.}
This research is supported by the National Research Foundation, Singapore under its AI Singapore Programmes (AISG Award No.: AISG2-RP-2021-021; AISG Award No.: AISG-100E2018-005). 
This project is also supported by SUTD project PIE-SGP-AI-2018-01.
We also gratefully acknowledge the support of NVIDIA AI Technology Center (NVAITC) for our research.

%% ----------------End of Discussion and Conclusion---------------------
\newpage
% \bibliography{icml2022}
% \bibliographystyle{icml2022}

%% ====================== Supplementary =========================
\newpage
% \quad
% \pagebreak

% ========== Contents ===========
% \pagebreak
\section*{Supplementary Materials}

\section*{Contents of this Supplementary}
This Supplementary provides additional experiments, results (penultimate layer visualization and $\eta$ analysis), case studies and analyses to further support our main finding on Systematic diffusion. The Supplementary materials are organized as follows:

\begin{itemize}
    \item Section \ref{sup-sec:additional-visualizations}: Additional Penultimate Layer Visualizations
    
    \item Section \ref{sup-sec:additional-exp}: Additional Experiments / Analysis
    
    % Reproducibility details
    \item Section \ref{sup-sec:reproducibility} : Research Reproducibility Details
    
    \item Section \ref{sup-sec:std-experiments}: Standard Deviation for main paper experiments
    
    \item Section \ref{sup-sec:additional-systematic-diffusion}: Additional Discussion: Why this diffusion is systematic and not isotopic?
    
    \item Section \ref{sup-sec:visualization-algorithm}: Algorithm for Projection and visualization of penultimate layer representations
    
    % Rebuttal additions
    \item Section \ref{sup-sec:semantic-classes}: Semantically similar / dissimilar classes
        \begin{itemize}
            \item Section \ref{sub-sec:sup-semantics-method1}: Using standard, pre-defined ImageNet knowledge graph as a prior
            
            \item Section \ref{sub-sec:sup-semantics-method2}: Using distance in the feature space
            
        \end{itemize}

    \item Section \ref{sup-sec:smoothness}: Case study: Smoothness of targets are insufficient to determine KD performance
        \begin{itemize}
            \item Section \ref{sub-sec:sup-case-study-1}: Case study at lower $T$ with same degree of smoothness
            
            \item Section \ref{sub-sec:sup-case-study-2}: Case study at moderately higher $T$ with same degree of smoothness
            
            \item Section \ref{sub-sec:sup-case-study-3}: Case study at very high $T$ with same degree of smoothness
            
        \end{itemize}
    
    \item Section \ref{sup-sec:class-wise-accuracy}: Class-wise accuracy for target classes

    \item Section 
    \ref{sup-sec:additional_alpha_T}:
Additional Exploration of $\alpha$ and $T$

        \item Section 
\ref{sup-sec:alternative_characterization}:
Alternative characterization of cluster distance

 \item Section 
\ref{sup-sec:sample_images}:
Sample images of different classes used in the study

\end{itemize}

% \clearpage

% =============Appendix==================
\appendix

\setcounter{figure}{0} 
\setcounter{table}{0} 
\renewcommand\thefigure{\thesection.\arabic{figure}}    
\renewcommand\thetable{\thesection.\arabic{table}}    

\newpage
% \quad
%\pagebreak
% \section{Visualizations and $\eta$ analysis for ResNet-50 Student (ImageNet-1K)}
% \label{sup-sec:r50-imagenet-analysis}

%%% ==================Additional Penultimate Layer Visualizations =============
\section{Additional Penultimate Layer Visualizations}
\label{sup-sec:additional-visualizations}
\setcounter{figure}{0} 
\setcounter{table}{0} 

In this section, we show additional penultimate layer visualizations to support our finding on systematic diffusion. The details are included in Table \ref{sup-table:additional-visualization-details}.

\begin{table}[h]
    \centering
    \caption{
    Penultimate layer visualization details supporting our finding on systematic diffusion.
    Our visualizations cover multiple tasks including image classification (standard, fine-grained) and compact student distillation tasks spanning across multiple datasets and teacher-student architectures.
    }
    
    \begin{adjustbox}{width=0.99\columnwidth,center}
       \begin{tabular}{lcc}\toprule
        \textbf{Teacher / Student} &\textbf{Dataset} &\textbf{Visualization} \\ \midrule
        ResNet-50 / ResNet-18 &ImageNet-1K &Figure \ref{fig:main} \\  \midrule
        ResNet-50 / ResNet-50 &ImageNet-1K &Figure \ref{sup-fig:r50-imagenet} \\ \midrule
        ResNet-50 / EfficientNet-B0 &ImageNet-1K &Figure \ref{sup-fig:efb0-imagenet} \\ \midrule
        ResNet-50 / ResNet-18 &CUB200 &Figure \ref{sup-fig:r18-cub} \\ \midrule
        ResNet-50 / ResNet-50 &CUB200 &Figure \ref{sup-fig:r50-cub} \\ \midrule
        ResNet-50 / ConvNext-T &CUB200 &Figure \ref{sup-fig:convnext-cub} \\
        \bottomrule
        \end{tabular}

    \end{adjustbox}
    
\label{sup-table:additional-visualization-details}
\end{table}

%%% ==================ResNet-50 Analysis (ImageNet-1k)=============
\begin{figure*}[ht]
\begin{adjustbox}{width=0.85\textwidth, center}
\begin{tabular}{c}
    \includegraphics[width=0.75\textwidth]{./pics_main/imagenet/legend.pdf} \\  
      \includegraphics[width=\textwidth]{./pics_main/imagenet/teacher_resnet50_ls_imagenet_m-poodle_s-poodle_submarine.pdf} \\   
      \includegraphics[width=\textwidth]{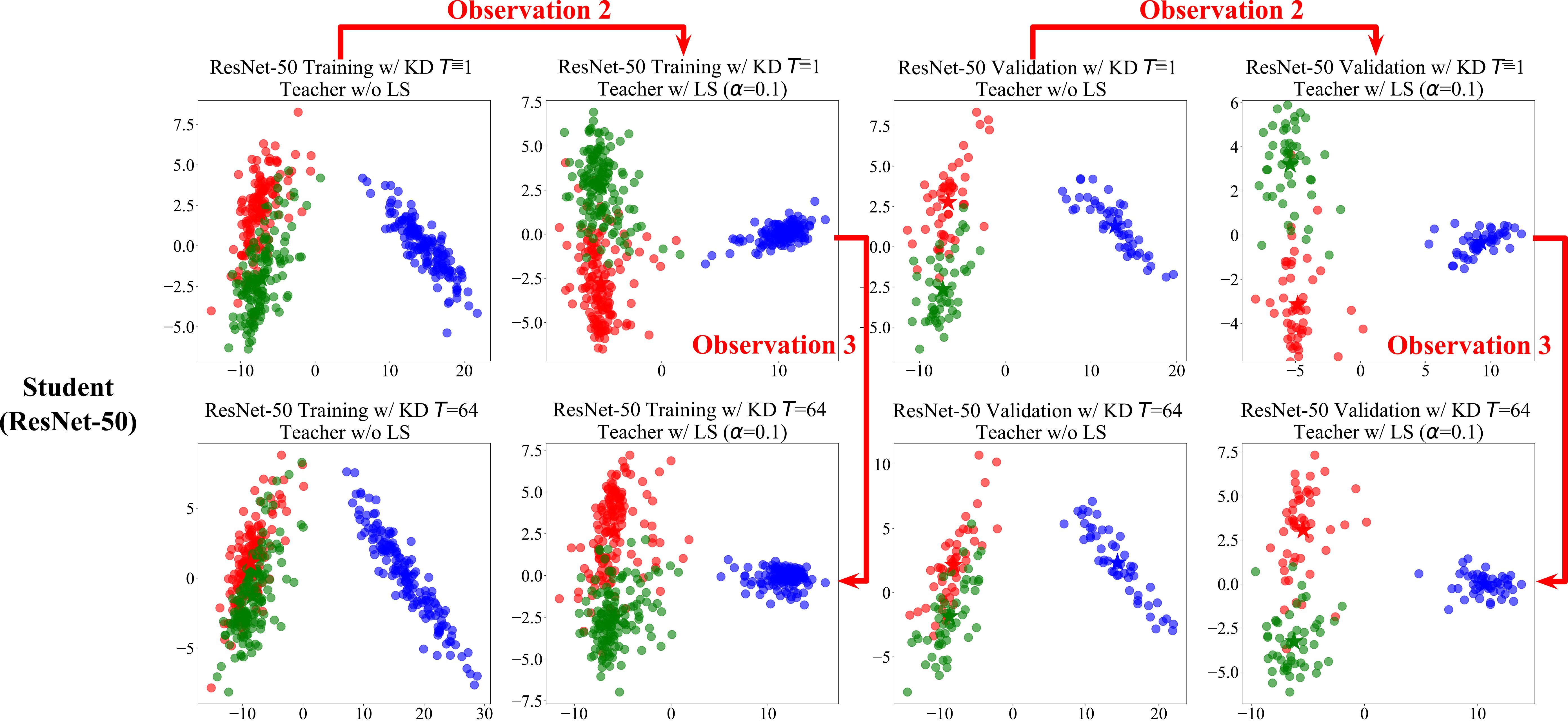} \\
\end{tabular}
\end{adjustbox}
\caption{Visualization of the penultimate layer representations
(\texttt{Teacher = ResNet-50, Student = ResNet-50, Dataset = ImageNet}).
We follow the same setup and procedure used in \citet{muller} and \citet{shen2021}. We also follow their three-class analysis: two semantically similar classes
({\tt \color{red} miniature\_poodle}, {\tt \color{green} standard\_poodle}) and one semantically different class ({\tt \color{blue} submarine}). {\color{red} \bf Observation 1:}
The use of LS on the teacher leads to tighter clusters and erasure of logits' information as claimed by \citet{muller}. In addition, increase in central distance between semantically similar classes ({\tt \color{red} miniature\_poodle},
{\tt \color{green} standard\_poodle}) as claimed by \citet{shen2021} can be observed.
{\color{red} \bf Observation 2:}
We further visualize the student's representations. 
Increase in central distance between semantically similar classes can also be observed. This confirms the transfer of this benefit from the teacher to the student. Note that in  
\citet{muller} and \citet{shen2021}, student's representations have not been visualized.
{\color{red} \bf Observation 3 (Our main discovery):} KD of an increased $T$ causes systematic diffusion of representations between semantically similar classes ({\tt \color{red} miniature\_poodle},
{\tt \color{green} standard\_poodle}). 
Since the student is also a very powerful network (ResNet-50), the extent of this systematic diffusion is not large compared to the ResNet-18 student. 
We further show $\eta$ analysis in Table \ref{table:eta} to quantitatively show this systematic diffusion. 
We also show image samples for these 3 classes in Figure \ref{sup-fig:samples_imagenet}.
Best viewed in color.}
\label{sup-fig:r50-imagenet}
%\vspace{-0.72cm}
\end{figure*}

%%% ==================EfficientNetB0 Analysis (ImageNet-1k)============
\begin{figure*}[ht]
\begin{adjustbox}{width=0.85\textwidth, center}
\begin{tabular}{c}
    \includegraphics[width=0.75\textwidth]{./pics_main/imagenet/legend.pdf} \\  
      \includegraphics[width=\textwidth]{./pics_main/imagenet/teacher_resnet50_ls_imagenet_m-poodle_s-poodle_submarine.pdf} \\   
      \includegraphics[width=\textwidth]{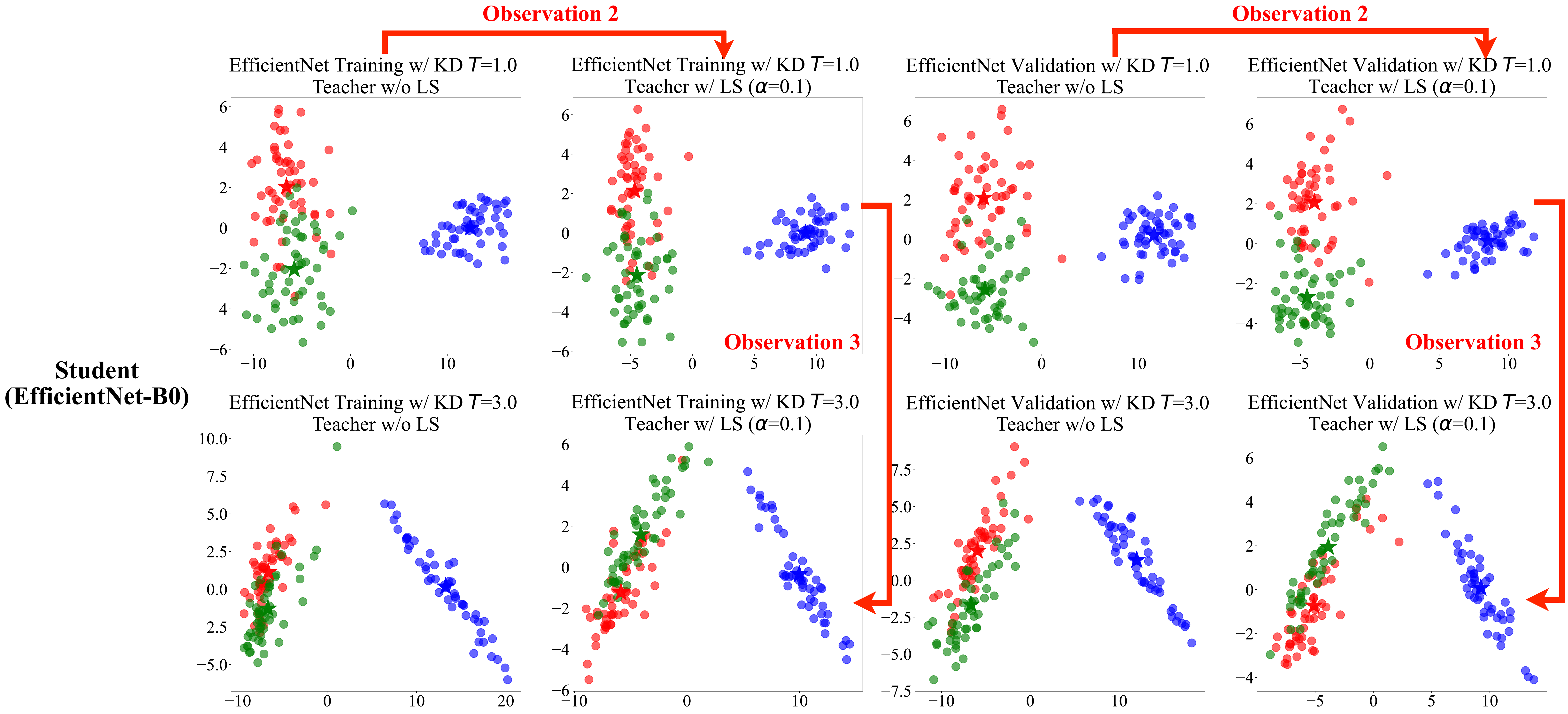} \\
\end{tabular}
\end{adjustbox}
\caption{Visualization of the penultimate layer representations
(\texttt{Teacher = ResNet-50, Student = EfficientNet-B0, Dataset = ImageNet}).
We follow the same setup and procedure used in \citet{muller} and \citet{shen2021}. We also follow their three-class analysis: two semantically similar classes
({\tt \color{red} miniature\_poodle}, {\tt \color{green} standard\_poodle}) and one semantically different class ({\tt \color{blue} submarine}). {\color{red} \bf Observation 1:}
The use of LS on the teacher leads to tighter clusters and erasure of logits' information as claimed by \citet{muller}. In addition, increase in central distance between semantically similar classes ({\tt \color{red} miniature\_poodle},
{\tt \color{green} standard\_poodle}) as claimed by \citet{shen2021} can be observed.
{\color{red} \bf Observation 2:}
We further visualize the student's representations. 
Increase in central distance between semantically similar classes can also be observed. This confirms the transfer of this benefit from the teacher to the student. Note that in  
\citet{muller} and \citet{shen2021}, student's representations have not been visualized.
{\color{red} \bf Observation 3 (Our main discovery):} KD of an increased $T$ causes systematic diffusion of representations between semantically similar classes ({\tt \color{red} miniature\_poodle},
{\tt \color{green} standard\_poodle}). 
We also show image samples for these 3 classes in Figure \ref{sup-fig:samples_imagenet}.
Best viewed in color.}
\label{sup-fig:efb0-imagenet}
%\vspace{-0.72cm}
\end{figure*}

\begin{figure*}[!ht]
\begin{adjustbox}{width=\textwidth,center}
\begin{tabular}{c}
    \includegraphics[width=0.75\textwidth]{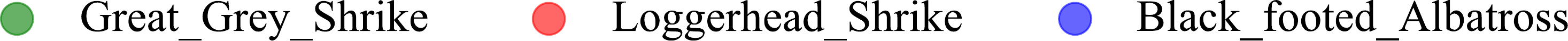} \\  
      \includegraphics[width=\textwidth]{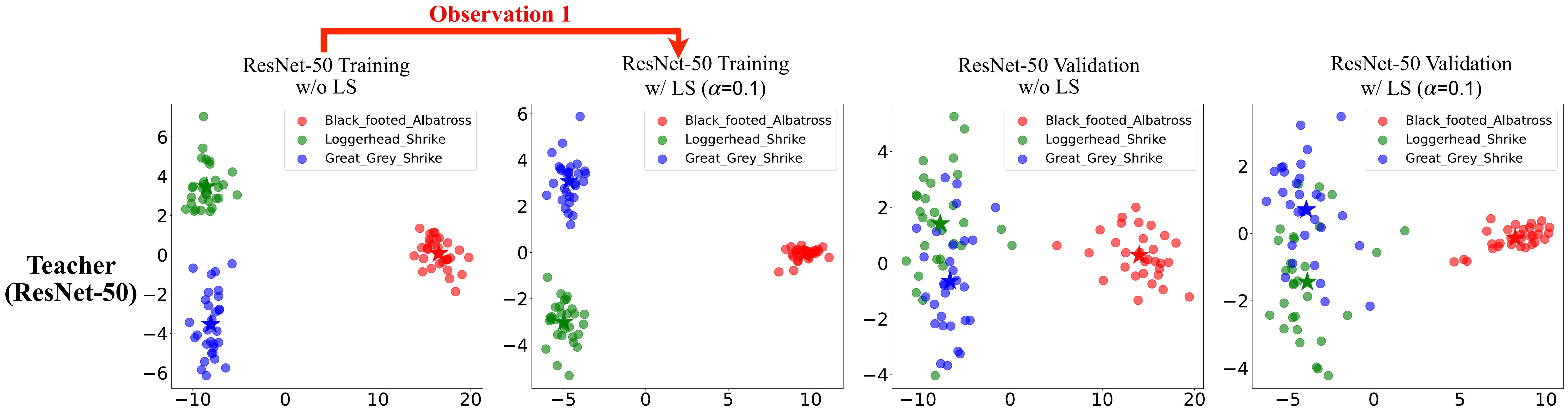} \\   
      \includegraphics[width=\textwidth]{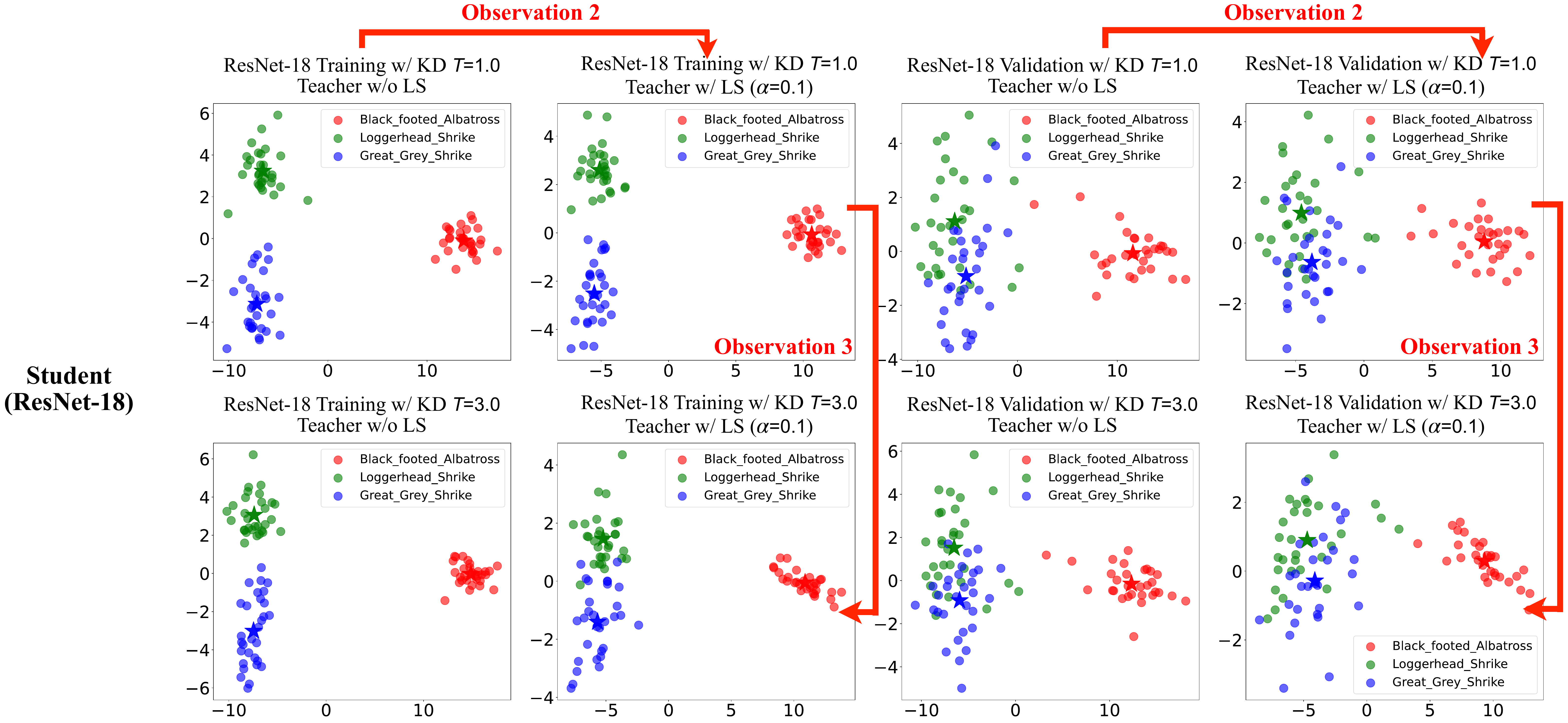} \\
\end{tabular}
\end{adjustbox}
\caption{Visualization of the penultimate layer representations
(\texttt{Teacher = ResNet-50, Student = ResNet-18, Dataset = CUB200-2011}).
We follow 
the same setup and procedure used in \citet{muller}
and 
\citet{shen2021}. We also follow their three-class analysis: two semantically similar classes
({\tt \color{red} Loggerhead\_Shrike}, {\tt \color{green} Great\_Grey\_Shrike})
and one semantically different class ({\tt \color{blue} Black\_footed\_Albatross}).
{\color{red} \bf Observation 1:}
The use of LS on the teacher leads to   
tighter clusters and erasure of logits' information as claimed by \citet{muller}. In addition, increase in central distance between semantically similar classes ({\tt \color{red} Loggerhead\_Shrike},
{\tt \color{green} Great\_Grey\_Shrike}) as claimed by \citet{shen2021} can be observed.
{\color{red} \bf Observation 2:}
We further visualize the student's representations. 
Increase in central distance between semantically similar classes can also be observed. This confirms the transfer of this benefit from the teacher to the student. Note that in  
\citet{muller} and \citet{shen2021}, student's representations have not been visualized.
{\color{red} \bf Observation 3 (Our main discovery):} KD of an increased $T$ causes systematic diffusion of representations between semantically similar classes ({\tt \color{red} Loggerhead\_Shrike},
{\tt \color{green} Great\_Grey\_Shrike}). 
We also show image samples for these 3 classes in Figure \ref{sup-fig:samples_cub}.
Best viewed in color.}
\label{sup-fig:r18-cub}
%\vspace{-0.72cm}
\end{figure*}

\begin{figure*}
\begin{adjustbox}{width=\textwidth,center}
\begin{tabular}{c}
    \includegraphics[width=0.75\textwidth]{./pics_main/cub/legend.pdf} \\  
      \includegraphics[width=\textwidth]{./pics_main/cub/teacher_resnet50_ls_CUB_Loggerhead_Shrike_Great_Grey_Shrike_Black_footed_Albatross.pdf} \\   
      \includegraphics[width=\textwidth]{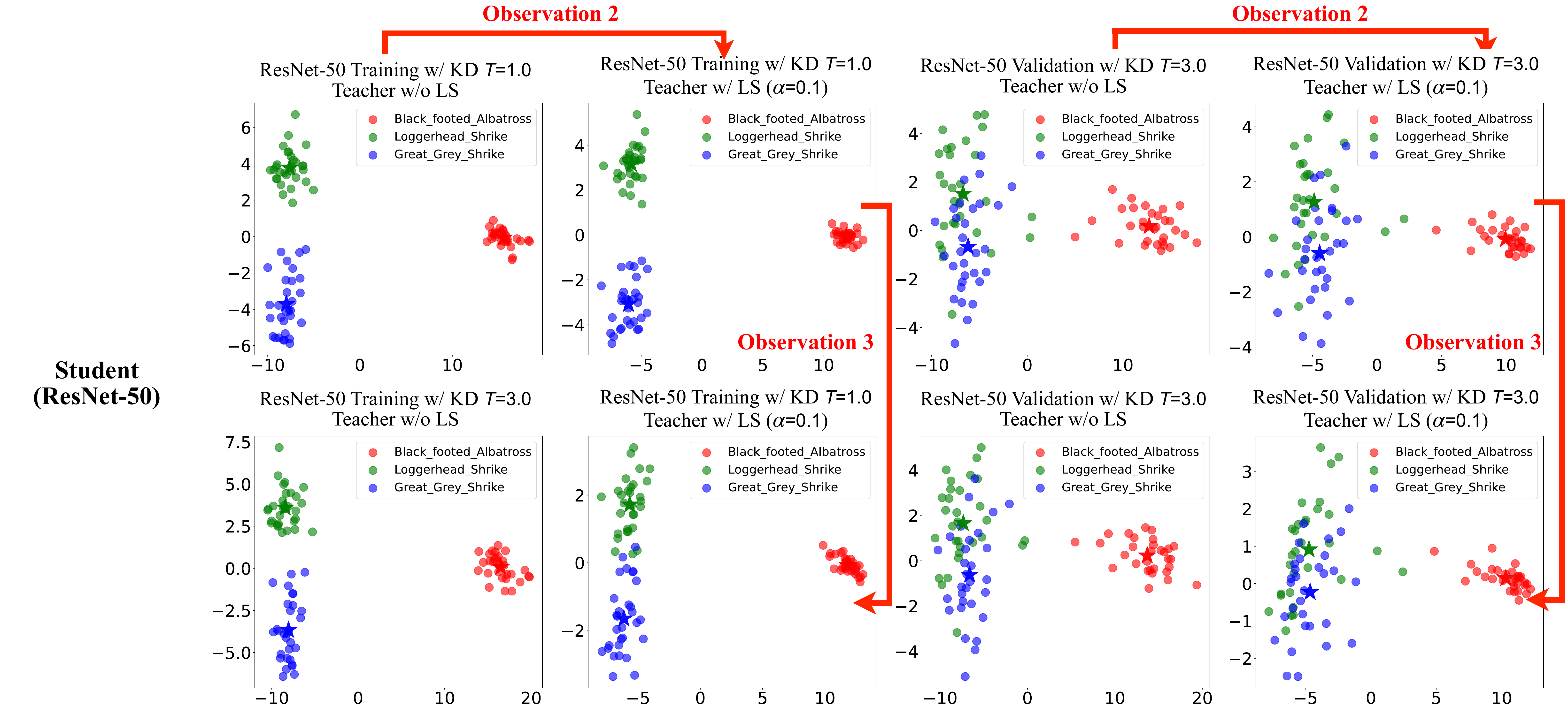} \\
\end{tabular}
\end{adjustbox}
\caption{Visualization of the penultimate layer representations
(\texttt{Teacher = ResNet-50, Student = ResNet-50, Dataset = CUB200-2011}).
We follow 
the same setup and procedure used in \citet{muller}
and 
\citet{shen2021}. We also follow their three-class analysis: two semantically similar classes
({\tt \color{red} Loggerhead\_Shrike}, {\tt \color{green} Great\_Grey\_Shrike})
and one semantically different class ({\tt \color{blue} Black\_footed\_Albatross}).
{\color{red} \bf Observation 1:}
The use of LS on the teacher leads to   
tighter clusters and erasure of logits' information as claimed by \citet{muller}. In addition, increase in central distance between semantically similar classes ({\tt \color{red} Loggerhead\_Shrike},
{\tt \color{green} Great\_Grey\_Shrike}) as claimed by \citet{shen2021} can be observed.
{\color{red} \bf Observation 2:}
We further visualize the student's representations. 
Increase in central distance between semantically similar classes can also be observed. This confirms the transfer of this benefit from the teacher to the student. Note that in  
\citet{muller} and \citet{shen2021}, student's representations have not been visualized.
{\color{red} \bf Observation 3 (Our main discovery):} KD of an increased $T$ causes systematic diffusion of representations between semantically similar classes ({\tt \color{red} Loggerhead\_Shrike},
{\tt \color{green} Great\_Grey\_Shrike}). 
We also show image samples for these 3 classes in Figure \ref{sup-fig:samples_cub}.
Best viewed in color.}
\label{sup-fig:r50-cub}
%\vspace{-0.72cm}
\end{figure*}

\begin{figure*}
\begin{adjustbox}{width=\textwidth,center}
\begin{tabular}{c}
    \includegraphics[width=0.75\textwidth]{./pics_main/cub/legend.pdf} \\  
      \includegraphics[width=\textwidth]{./pics_main/cub/teacher_resnet50_ls_CUB_Loggerhead_Shrike_Great_Grey_Shrike_Black_footed_Albatross.pdf} \\   
      \includegraphics[width=\textwidth]{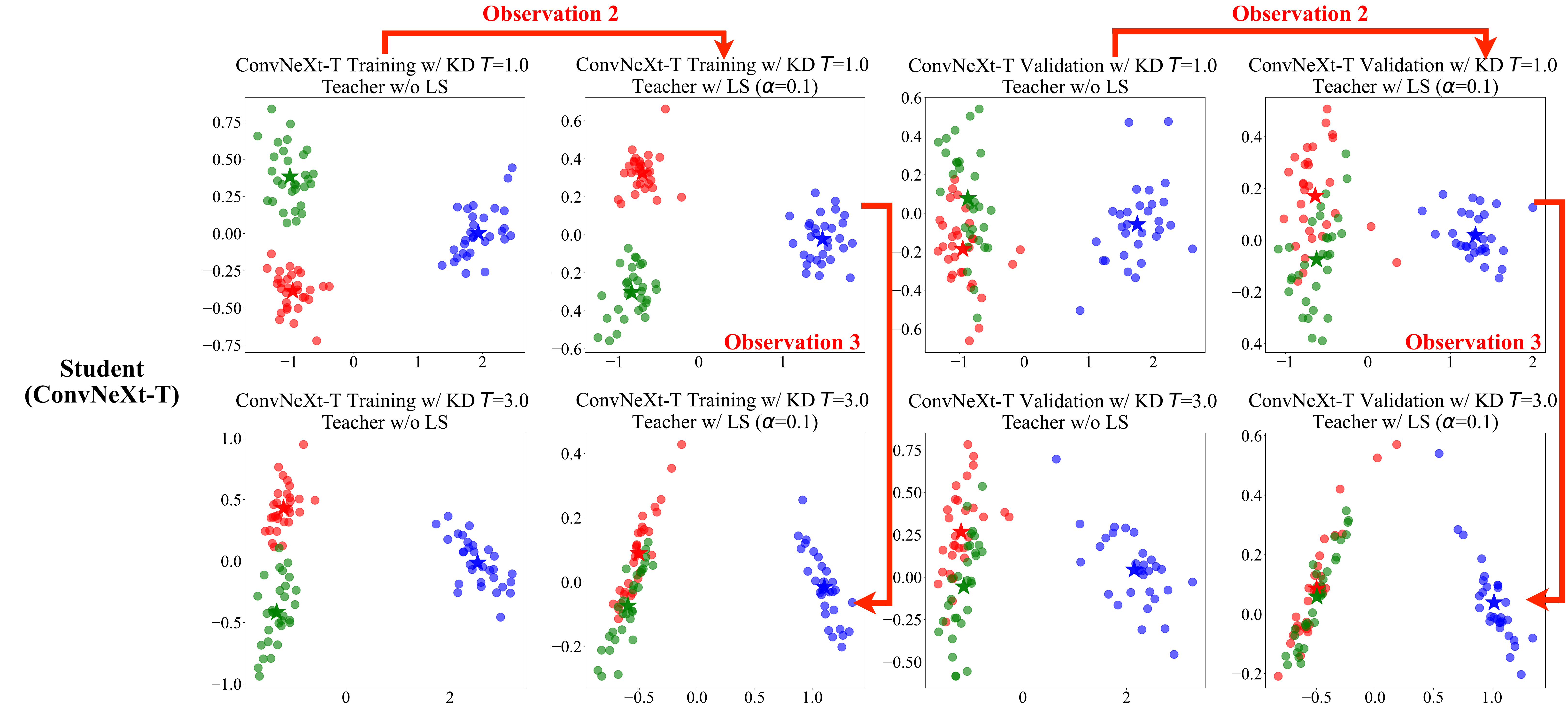} \\
\end{tabular}
\end{adjustbox}
\caption{Visualization of the penultimate layer representations
(\texttt{Teacher = ResNet-50, Student = ConvNeXt-T, Dataset = CUB200-2011}).
We follow 
the same setup and procedure used in \citet{muller}
and 
\citet{shen2021}. We also follow their three-class analysis: two semantically similar classes
({\tt \color{red} Loggerhead\_Shrike}, {\tt \color{green} Great\_Grey\_Shrike})
and one semantically different class ({\tt \color{blue} Black\_footed\_Albatross}).
{\color{red} \bf Observation 1:}
The use of LS on the teacher leads to   
tighter clusters and erasure of logits' information as claimed by \citet{muller}. In addition, increase in central distance between semantically similar classes ({\tt \color{red} Loggerhead\_Shrike},
{\tt \color{green} Great\_Grey\_Shrike}) as claimed by \citet{shen2021} can be observed.
{\color{red} \bf Observation 2:}
We further visualize the student's representations. 
Increase in central distance between semantically similar classes can also be observed. This confirms the transfer of this benefit from the teacher to the student. Note that in  
\citet{muller} and \citet{shen2021}, student's representations have not been visualized.
{\color{red} \bf Observation 3 (Our main discovery):} KD of an increased $T$ causes systematic diffusion of representations between semantically similar classes ({\tt \color{red} Loggerhead\_Shrike},
{\tt \color{green} Great\_Grey\_Shrike}). 
We also show image samples for these 3 classes in Figure \ref{sup-fig:samples_cub}.
Best viewed in color.}
\label{sup-fig:convnext-cub}
%\vspace{-0.72cm}
\end{figure*}

%%% ==================End of Additional Penultimate Layer Visualizations =============

%%% ==================Extended Experiments =============
%\clearpage
\section{Additional Experiments / Analysis}
\label{sup-sec:additional-exp}
\setcounter{figure}{0} 
\setcounter{table}{0} 

\textbf{Fine-grained image classification.}
We conduct experiments using an additional student architecture, ConvNeXt-T \cite{liu2022convnet} to further support our findings. The results are shown in Table \ref{sup-table:convnext-cub}. 
We show systematic diffusion using penultimate layer visualizations in Figure \ref{sup-fig:convnext-cub}.

\begin{table}[h]
    \centering
    \caption{
    Top1 / Top5 accuracies for fine-grained classification (CUB200) using ConvNeXt-T student.
    We use $T=1, T=3$ for distilling knowledge from ResNet-50 teacher.
    As one can clearly observe, \emph{with LS-trained teacher, there is a consistent degrade in student performance as $T$ increases. This can be observed in all our 34 experiments.}
    These results comprehensively support our claim: \emph{in the presence of an LS-trained teacher, KD at higher temperatures is rendered ineffective.}
    }
    \begin{adjustbox}{width=\columnwidth,center}
        \begin{tabular}{c|c|c|c} \toprule
        &\backslashbox{T}{$\alpha$} &$\alpha=0.0$ &$\alpha=0.1$ \\ \midrule
        Teacher : ResNet-50 &- &81.584 / 95.927 &82.068 / 96.168 \\ \midrule
        Student : &$T$ = 1 &86.624 / 97.221 &\textbf{86.866 / 97.377} \\ \cmidrule{2-4}
        ConvNeXt-T
        &$T$ = 3 &86.554 / 97.187 &83.638 / 97.135 \\
        \bottomrule
        \end{tabular}

    \end{adjustbox}
    
    \label{sup-table:convnext-cub}
\end{table}

\textbf{Neural Machine Translation.} 
We conduct additional KD experiments for English $\rightarrow$ Russian translation using IWSLT dataset. 
We report BLUE scores in Table \ref{sup-table:nmt-en-ru}.
We remark that all these experiments comprehensively support our main finding on systematic diffusion.

\begin{table}[h]
    \centering
    \caption{
    BLEU scores for KD experiments from Transformer Teacher to Transformer student on IWSLT dataset using English $\rightarrow$ Russian translation task, following the similar procedure as \citet{shen2021}. 
    Configurations where LS and KD are compatible are in \textbf{bold}.
    As one can clearly observe, \emph{with LS-trained teacher, there is a consistent degrade in student performance as $T$ increases. This can be observed in all our 34 experiments.}
    These results comprehensively support our claim: \emph{in the presence of an LS-trained teacher, KD at higher temperatures is rendered ineffective.}
    }
        
    \begin{adjustbox}{width=0.35\textwidth,center}
        \begin{tabular}{l|c|c|c}\toprule
        &\backslashbox{T}{$\alpha$} &$\alpha$ = 0.0 &$\alpha$ = 0.1 \\ \toprule
        Teacher : Transformer &- &16.718 &16.976 \\ \midrule
        \multirow{4}{*}{Student : Transformer} &$T$ = 1 &16.140 &\textbf{16.197} \\ \cmidrule{2-4}
        &$T$ = 2 &14.977 &\textbf{15.100} \\ \cmidrule{2-4}
        &$T$ = 3 &13.826 &\textbf{14.106} \\ \cmidrule{2-4}
        &$T$ = 64 &3.605 &3.590 \\
        \bottomrule
        \end{tabular}

    \end{adjustbox}
    
\label{sup-table:nmt-en-ru}
\end{table}

\textbf{Compact Student Distillation.}
We conduct experiments using an additional compact student architecture, EfficientNet-B0 \cite{tan2019efficientnet} (5.3M params) to further support our findings. We use ResNet-50 as the teacher model and perform large-scale experiments using ImageNet-1K. 
The results are shown in Table \ref{sup-table:efb0-imagenet}. 
We show systematic diffusion using penultimate layer visualizations in Figure \ref{sup-fig:efb0-imagenet} and $\eta$ results are shown in Table \ref{sup-table:efb0-imagenet-eta}.

\begin{table}[h]
    \centering
    \caption{
    Top1 / Top5 accuracies for compact student distillation (ImageNet-1K) using EfficientNet-B0 student.
    We use $T=1, T=3$ for distilling knowledge from ResNet-50 teacher.
    As one can clearly observe, \emph{with LS-trained teacher, there is a consistent degrade in student performance as $T$ increases. This can be observed in all our 34 experiments.}
    These results comprehensively support our claim: \emph{in the presence of an LS-trained teacher, KD at higher temperatures is rendered ineffective.}
    }
    \begin{adjustbox}{width=\columnwidth,center}
        \begin{tabular}{c|c|c|c} \toprule
        &\backslashbox{T}{$\alpha$} &$\alpha=0.0$ &$\alpha=0.1$ \\  \midrule
        Teacher : ResNet-50 &- &76.130 / 92.862 &76.196 / 93.078 \\  \midrule
        Student : &$T$ = 1 &68.850 / 88.604 &\textbf{69.906 / 89.284} \\ \cmidrule{2-4}
        EfficientNet-B0
        &$T$ = 3 &68.546 / 88.704 &58.182 / 83.918 \\
        \bottomrule
        \end{tabular}

    \end{adjustbox}
    \label{sup-table:efb0-imagenet}
\end{table}

\begin{table}[h]
    \caption{
    $\eta$ calculation for EfficientNet-B0 for 10 target classes (exact classes used in Table \ref{table:eta} main paper).
    Our finding is consistently observed (see main paper).
    We clearly show that $\eta(T_1=1, T_2=3; \pi, S_1) < 0$ and $\eta(T_1=1, T_2=3; \pi, S_2) > 0$ for all these 10 target classes, thereby quantitatively 
    demonstrating our discovery on systematic diffusion.
    }
    \begin{adjustbox}{width=0.95\columnwidth,center}
      \begin{tabular}{lcccc}
        \multicolumn{5}{c}{\large{Set 1}}\\
        \toprule
        Target class &$Train:S_1$ &$Train:S_2$ &$Val:S_1$ &$Val:S_2$ \\ \toprule
        Chesapeake\_Bay\_retriever &-2.276 &0.490 &-3.760 &0.790 \\ \midrule
        curly-coated\_retriever &-0.830 &0.235 &-4.502 &0.933 \\ \midrule
        flat-coated\_retriever &-0.979 &0.173 &-3.904 &0.807 \\ \midrule
        golden\_retriever &-3.651 &0.694 &-4.356 &0.890 \\ \midrule
        Labrado\_retriever &-2.747 &0.469 &-4.730 &0.860 \\
        \bottomrule
        \vspace{0.1cm}
        \end{tabular}
    \end{adjustbox}
    
    \begin{adjustbox}{width=0.90\columnwidth,center}
       \begin{tabular}{lcccc}
        \multicolumn{5}{c}{\large{Set 2}}\\
        \toprule
        Target class &$Train:S_1$ &$Train:S_2$ &$Val:S_1$ &$Val:S_2$ \\ \toprule
        thunder\_snake &-10.981 &1.458 &-12.916 &1.789 \\  \midrule
        ringneck\_snake &-9.629 &1.211 &-10.617 &1.373 \\  \midrule
        hognose\_snake &-7.984 &1.271 &-9.347 &1.536 \\  \midrule
        water\_snake &-8.217 &1.302 &-9.645 &1.489 \\  \midrule
        king\_snake &-8.371 &1.365 &-10.082 &1.647 \\
        \bottomrule
        \end{tabular}
    
    \end{adjustbox}
\label{sup-table:efb0-imagenet-eta}
\end{table}

\textbf{Advanced KD methods.}
We demonstrate our finding using a popular advanced KD method by \citet{heo2019comprehensive}.
This method performs optimized feature distillation
(contains margin-ReLU feature transforms, partial L2 distance functions, optimized feature positions)
combined with task loss (In our experiments, the task is KD)
\renewcommand{\footnotesize}{\tiny} 
\footnote{\url{https://github.com/clovaai/overhaul-distillation}}
.
\renewcommand{\footnotesize}{\small} 
The results are shown in Table \ref{sup-table:convnext-advanced-kd}.
We show that advanced KD methods also suffer from systematic diffusion when distilling from an LS-trained teacher at larger $T$.

\begin{table}[h]
    \centering
    \caption{
    Advanced KD results using method proposed by \cite{heo2019comprehensive}: We show top1 / top5 accuracies for fine-grained classification (CUB200) using ResNet-50 teacher to ResNet-50 student.
    We use $T=1, T=3$ for distilling knowledge from ResNet-50 teacher.
    As one can clearly observe, \emph{with LS-trained teacher, there is a consistent degrade in student performance as $T$ increases when using advanced KD methods.}
    These results comprehensively support our claim: \emph{in the presence of an LS-trained teacher, KD at higher temperatures is rendered ineffective.}
    }
    \begin{adjustbox}{width=\columnwidth,center}
        \begin{tabular}{c|c|c|c} \toprule
        &\backslashbox{T}{$\alpha$} &$\alpha=0.0$ &$\alpha=0.1$ \\ \midrule
        Teacher : ResNet-50 &- &81.584 / 95.927 &82.068 / 96.168 \\ \midrule
        Student : ResNet-50 &$T$ = 1 &82.568 / 96.479 &\textbf{83.794 / 96.686} \\ \cmidrule{2-4}
        Advanced KD
        &$T$ = 3 &82.706 / 96.307 &81.739 / 96.117 \\
        \bottomrule
        \end{tabular}

    \end{adjustbox}
    
    \label{sup-table:convnext-advanced-kd}
\end{table}

%% ----------------Reproducibility Statement---------------------
\section{Research Reproducibility Details}
\label{sup-sec:reproducibility}

\textbf{Code / Pre-trained models.}
Pytorch code to reproduce all our results and analysis can be found at \url{https://keshik6.github.io/revisiting-ls-kd-compatibility/}. 
All pretrained models for image classification using ImageNet-1K, fine-grained classification using CUB200-2011, neural machine translation using IWSLT and compact student distillation are available at \url{https://keshik6.github.io/revisiting-ls-kd-compatibility/}. 

\textbf{Docker information :} To allow for training in containerised environments (HPC, Super-computing clusters), please use \textit{nvcr.io/nvidia/pytorch:20.12-py3} container.

\textbf{Experiment details and hyper-parameters:}

\textit{ImageNet-1K}: For ImageNet experiments, we follow similar setup as \citet{shen2021} and use ILSVRC2012 version. 
For training LS networks, we train for 90 epochs with initial learning rate 0.1 decayed by a factor of 10 every 30 epochs. 
For KD experiments, we train for 200 epochs with initial learning rate 0.1 decayed by a factor of 10 every 80 epochs. 
We conducted a grid search for hyper-parameters as well.
For all experiments, we use a batch size of 256 and SGD with momentum 0.9 .
For data augmentation, we use random crops and random horizontal flips. 
All experiments were repeated 3 times.
For visualization of penultimate layer representations, we use 150 samples for training set and 50 samples for validation set.

\textit{Fine-grained classification and compact student distillation. } We follow similar setup as \citet{shen2021}.
For training both LS and KD networks, we train for 200 epochs with initial learning rate 0.01 decayed by a factor of 10 every 80 epochs. 
We conducted a grid search for hyper-parameters as well.
For all experiments, we use a batch size of 256 and SGD with momentum 0.9 .
All experiments were repeated 3 times.
For data augmentation, we use random crops, random rotation, color jitter and random horizontal flips.
For visualization of penultimate layer representations, we use all samples for training and validation sets.

\textit{Neural Machine Translation } 
We use IWSLT dataset.
We follow similar setup as \citet{shen2021}.
We use Adam as the optimizer, lr with 0.0005, dropout
with drop rate as 0.3, weight-decay with 0 and max tokens with 4096, all of these hyper-parameters
are following settings of \citet{shen2021}.
These hyper-parameters were used for both translation tasks (English $\rightarrow$ German, English $\rightarrow$ Russian). We use the code \href{https://github.com/RayeRen/multilingual-kd-pytorch}{here} similar to \citet{shen2021}.
%% ----------------End of Reproducibility Statement---------------------

%% -------------- Standard deviation of ImageNet-1K experiments ---------------
% \clearpage
\section{Standard Deviation for main paper experiments}
\label{sup-sec:std-experiments}
\setcounter{figure}{0} 
\setcounter{table}{0} 

In this section, we report the standard deviation for all KD experiments in the main paper. The standard deviation for ImageNet-1K and CUB200-2011 main paper experiments are reported in Tables \ref{table:sup-std-imagenet} and \ref{table:sup-std-cub} respectively.
The standard deviation for Compact student distillation and neural machine translation main paper experiments are reported in Tables \ref{sup-table:mobilenetv2-cub-std} and \ref{table:nmt-en-de-std} respectively.
All standard deviations are within acceptable range.

% For ImageNet-1K
\begin{table*}[h]
    \centering
    \caption{
    KD results from ResNet-50 Teacher to ResNet-18, ResNet-50 students \textbf{with standard deviations}, following similar procedure as \citet{shen2021} on ImageNet-1K \citep{imagenet_cvpr09}. 
    We show the top1/ top5 test accuracies. 
    Configurations where LS and KD are compatible are in \textbf{bold}.
    As one can clearly observe, \emph{with LS-trained teacher, there is a consistent degrade in student performance as $T$ increases. This can be observed in all our 34 experiments.}
    These results comprehensively support our claim: \emph{in the presence of an LS-trained teacher, KD at higher temperatures is rendered ineffective.}
    On the other hand, we observe that higher $T$ can improve the performance when using a teacher trained without LS in fine-grained classification and compact student  distillation experiments (See Table \ref{table:results-imagenet-cub} (B) and Table \ref{table:mobilenetv2-cub})
    All these results are averaged over 3 independent runs. The standard deviations are well within acceptable range.
    }
        
    \begin{adjustbox}{width=0.80\textwidth,center}
        \begin{tabular}{l|c|c|c}\toprule
        &\backslashbox{$T$}{$\alpha$} &$\alpha$ = 0.0 &$\alpha$ = 0.1 \\ \toprule
        \multirow{4}{*}{Student : ResNet-18} &$T$ = 1 &71.547 $\pm$ 0.122 / 90.297 $\pm$ 0.175 &\textbf{71.616 $\pm$ 0.114 / 90.233 $\pm$ 0.119} \\ \cmidrule{2-4}
        &$T$ = 2 &71.349 $\pm$ 0.017 / 90.359 $\pm$ 0.054 &68.799 $\pm$ 0.065 / 89.279 $\pm$ 0.092 \\ \cmidrule{2-4}
        &$T$ = 3 &69.570 $\pm$ 0.320 / 89.657 $\pm$ 0.041 &67.699 $\pm$ 0.079 / 89.043 $\pm$ 0.096 \\ \cmidrule{2-4}
        &$T$ = 64 &66.230 $\pm$ 0.036 / 88.730 $\pm$ 0.071 &64.506 $\pm$ 0.142 / 87.811 $\pm$ 0.100 \\
        \midrule
        \multirow{4}{*}{Student : ResNet-50} &$T$ = 1 &76.502 $\pm$ 0.234 / 93.059 $\pm$ 0.061 &\textbf{77.035 $\pm$ 0.061 / 93.327 $\pm$ 0.185} \\ \cmidrule{2-4}
        &$T$ = 2 &76.198 $\pm$ 0.035 / 92.987 $\pm$ 0.105 &76.101 $\pm$ 0.105 / 93.115 $\pm$ 0.017 \\ \cmidrule{2-4}
        &$T$ = 3 &75.388 $\pm$ 0.095 / 92.676 $\pm$ 0.006 &\textbf{75.821 $\pm$ 0.006 / 93.065 $\pm$ 0.088} \\ \cmidrule{2-4}
        &$T$ = 64 &74.291 $\pm$ 0.014 / 92.399 $\pm$ 0.035 &\textbf{74.627 $\pm$ 0.035 / 92.639 $\pm$ 0.085} \\
        \bottomrule
        \end{tabular}

    \end{adjustbox}
    
\label{table:sup-std-imagenet}
\end{table*}

% For CUB
\begin{table*}[h]
    \centering
    \caption{
    KD results from ResNet-50 Teacher to ResNet-18, ResNet-50 students \textbf{with standard deviations}, following similar procedure as \citet{shen2021} on CUB200-2011 \citep{WahCUB_200_2011}. 
    We report top1/ top5 test accuracies. 
    Configurations where LS and KD are compatible are in \textbf{bold}.
    As one can clearly observe, \emph{with LS-trained teacher, there is a consistent degrade in student performance as $T$ increases. This can be observed in all our 34 experiments.}
    These results comprehensively support our claim: \emph{in the presence of an LS-trained teacher, KD at higher temperatures is rendered ineffective.}
    On the other hand, we observe that higher $T$ can improve the performance when using a teacher trained without LS in fine-grained classification and compact student  distillation experiments (See Table \ref{table:results-imagenet-cub} and Table \ref{table:mobilenetv2-cub}).
    These experiments are repeated for 3 independent runs and as you can observe the standard deviations are within acceptable range.
    }
        
    \begin{adjustbox}{width=0.80\textwidth,center}
        \begin{tabular}{l|c|c|c}\toprule
        &\backslashbox{$T$}{$\alpha$} &$\alpha$ = 0 &$\alpha$ = 0.1 \\ \toprule
        % Teacher : ResNet-50 &- &81.584 / 95.927 &82.068 / 96.168 \\ \midrule
      \multirow{4}{*}{Student : ResNet-18} &$T$ = 1 &80.169 $\pm$ 0.336 / 95.392 $\pm$ 0.03 &\textbf{80.946 $\pm$ 0.03 / 95.312 $\pm$ 0.18} \\ \cmidrule{2-4}
        &$T$ = 2 &80.808 $\pm$ 0.314 / 95.593 $\pm$ 0.053 &80.428 $\pm$ 0.053 / 95.518 $\pm$ 0.108 \\ \cmidrule{2-4}
        &$T$ = 3 &80.785 $\pm$ 0.26 / 95.674 $\pm$ 0.163 &78.196 $\pm$ 0.163 / 95.213 $\pm$ 0.125 \\ \cmidrule{2-4}
        &$T$ = 64 &73.611 $\pm$ 0.314 / 94.529 $\pm$ 0.086 &67.161 $\pm$ 0.086 / 93.062 $\pm$ 0.127 \\
        \midrule
        \multirow{4}{*}{Student : ResNet-50} &$T$ = 1 &82.902 $\pm$ 0.343 / 96.358 $\pm$ 0.141 &\textbf{83.742 $\pm$ 0.141 / 96.778 $\pm$ 0.12} \\ \cmidrule{2-4}
        &$T$ = 2 &82.534 $\pm$ 0.137 / 96.427 $\pm$ 0.105 &\textbf{83.379 $\pm$ 0.105 / 96.537 $\pm$ 0.018} \\ \cmidrule{2-4}
        &$T$ = 3 &82.091 $\pm$ 0.161 / 96.243 $\pm$ 0.13 &\textbf{82.142 $\pm$ 0.13 / 96.427 $\pm$ 0.211} \\ \cmidrule{2-4}
        &$T$ = 64 &79.784 $\pm$ 0.26 / 95.927 $\pm$ 0.13 &77.206 $\pm$ 0.13 / 95.812 $\pm$ 0.259 \\
        \bottomrule
        \end{tabular}

    \end{adjustbox}
    
\label{table:sup-std-cub}
\end{table*}

% NMT
\begin{table}
    \centering
    \caption{
    BLEU scores for KD experiments \textbf{with standard deviations} for Transformer Teacher to Transformer student on IWSLT dataset using English $\rightarrow$ German translation task, following the similar procedure as \citet{shen2021}. 
    Configurations where LS and KD are compatible are in \textbf{bold}.
    These results comprehensively support our claim: \emph{in the presence of an LS-trained teacher, KD at higher temperatures is rendered ineffective.}
    These experiments are repeated for 3 independent runs and standard deviations are within acceptable range.
    }
        
     \begin{adjustbox}{width=0.50\textwidth,center}
        \begin{tabular}{l|c|c|c}\toprule
        &\backslashbox{$T$}{$\alpha$} &$\alpha$ = 0.0 &$\alpha$ = 0.1 \\ \toprule
        % Teacher : Transformer &- &26.461 &26.750 \\ \midrule
        \multirow{4}{*}{Student : Transformer} &$T$ = 1 &24.914 $\pm$ 0.013 &\textbf{25.085 $\pm$ 0.082} \\ \cmidrule{2-4}
        &$T$ = 2 &23.103 $\pm$ 0.103 &\textbf{23.421 $\pm$ 0.039} \\ \cmidrule{2-4}
        &$T$ = 3 &21.999 $\pm$ 0.06 &\textbf{22.076 $\pm$ 0.125} \\ \cmidrule{2-4}
        &$T$ = 64 &6.564 $\pm$ 0.288 &6.461 $\pm$ 0.061 \\
        \bottomrule
        \end{tabular}

    \end{adjustbox}
    
\label{table:nmt-en-de-std}
\end{table}

% Compact DNN
\begin{table*}[h]
    \centering
    \caption{
    KD results \textbf{with standard deviations} from ResNet-50 Teacher to MobileNet-V2 (Compact DNN) student using CUB200
    Configurations where LS and KD are compatible are in \textbf{bold}.
    As one can clearly observe, \emph{with LS-trained teacher, there is a consistent degrade in student performance as $T$ increases. This can be observed in all our 34 experiments.}
    These results comprehensively support our claim: \emph{in the presence of an LS-trained teacher, KD at higher temperatures is rendered ineffective.}
    These experiments are repeated for 2 independent runs and as you can observe the standard deviations are within acceptable range.
    }
    
    \begin{adjustbox}{width=0.80\textwidth,center}
        \begin{tabular}{l|c|c|c}\toprule
        &\backslashbox{$T$}{$\alpha$} &$\alpha$ = 0.0 &$\alpha$ = 0.1 \\ \toprule
        % Teacher : ResNet-50 &- &81.584 / 95.927 &82.068 / 96.168 \\ \midrule
        \multirow{4}{*}{Student : ResNet-18} &$T$ = 1 &81.144 $\pm$ 0.037 / 95.677 $\pm$ 0.062 &\textbf{81.731 $\pm$ 0.256 / 95.754 $\pm$ 0.098} \\ \cmidrule{2-4}
        &$T$ = 2 &81.895 $\pm$ 0.024 / 95.858 $\pm$ 0.000 &80.609 $\pm$ 0.061 / 95.47 $\pm$ 0.159 \\ \cmidrule{2-4}
        &$T$ = 3 &81.257 $\pm$ 0.073 / 95.677 $\pm$ 0.012 &78.961 $\pm$ 0.293 / 95.306 $\pm$ 0.196 \\ \cmidrule{2-4}
        &$T$ = 64 &75.441 $\pm$ 0.049 / 94.702 $\pm$ 0.025 &70.435 $\pm$ 0.171 / 93.494 $\pm$ 0.025 \\
        \bottomrule
        \end{tabular}
    \end{adjustbox}
\label{sup-table:mobilenetv2-cub-std}
\end{table*}

%% -------------- End of Standard deviation of ImageNet experiments ---------------

%% -------------- Additional Discussion on why this diffusion is systematic? ---------------
% \clearpage
\section{Additional Discussion: Why this diffusion is systematic and not isotopic?}
\label{sup-sec:additional-systematic-diffusion}
\setcounter{figure}{0} 
\setcounter{table}{0} 

We provide more perspective into why this diffusion is systematic and not isotopic. 
We use the LS-trained ResNet-50 teacher (same one in Figure \ref{figure_2_logit}) trained on ImageNet-1K to numerically show more evidence as to why this diffusion is systematic and not isotopic.
Particularly we show that only very few classes (out of the 1000 classes in ImageNet-1K) have probabilities significantly larger than others. 
We examine the output probability for 3 classes: standard\_poodle samples, golden\_retriever samples and thunder\_snake samples (We choose this classes randomly, similar analysis can be done for other classes as well).

For each class, we compute the average output probability for 1300 training samples, and observe following:
Let $p_1$ be the largest probability which is also probability of the correct class.

\begin{itemize}[leftmargin=2em]
    \item For the average probability of standard\_poodle samples, the second largest probability, $p_2$ (miniature\_poodle) is at least 100x larger than 976 other probabilities (out of 999 probabilities)
    
    \item For the average probability of golden\_retriever samples, the second largest probability, $p_2$ (Labrador\_retriever) is at least 100x larger than 924 other probabilities (out of 999 probabilities) 
    
    \item For the average probability of thunder\_snake samples, the second largest probability, $p_2$ (ringneck\_snake) is at least 100x larger than 964 other probabilities (out of 999 probabilities) 
\end{itemize}

\textbf{Can this support the diffusion is systematic?}
We use results of standard\_poodle for discussion.
When KD of an increased T is used, these probabilities are scaled, and $p_2$ is brought closer to $p_1$, see Figure \ref{figure_2_logit}.
Consequently, student is encouraged to produce penultimate layer representations of standard\_poodle samples that are closer to miniature\_poodle.
This results in diffusion of penultimate layer representations of standard\_poodle towards  miniature\_poodle, curtailing the distance enlargement benefit of distilling from an LS-trained teacher.
For the 976 classes which have probabilities at least 100x smaller than that of miniature\_poodle, even with T scaling, the probabilities remain negligible. They have no influence on the representation of standard\_poodle.
Therefore diffusion of standard\_poodle will be towards miniature\_poodle and several semantically similar classes but there is no diffusion towards these 976 classes.
\textit{Therefore, the diffusion is systematic and is not isotopic.}

In this discussion, we use 100x to mean significance/insignificance. If a probability $p_i$ is 100x smaller than another probability $p_j$, then even with $T$ scaling $p_i$ remains insignificant compared to $p_j$.

%% -------------- End of Additional Discussion on why this diffusion is systematic? ---------------

%% -------------- Visualization algorithm ---------------
% \clearpage
\section{Algorithm for Projection and visualization of penultimate layer representations}
\setcounter{figure}{0} 
\setcounter{table}{0} 
The algorithm for projection and visualization is included in  \ref{sup-algorithm} \citet{muller}.
We also include a numpy style code of the projection / visualization algorithm in \ref{sup-algorithm-np}.

\label{sup-sec:visualization-algorithm}
\begin{algorithm*}[ht]
\caption{Projection and visualization of penultimate layer features} 
\begin{flushleft}

{\bf Input:}
\textcircled{\raisebox{-1pt}{1}} 
 High dimensional ($h$) features  $(X, Y)$ of three classes extracted from penultimate layers of the trained model $f$
\textcircled{\raisebox{-1pt}{2}} Model weight $w$ of the final layer of $f$
\\

{\bf Output: } The projected 2-D features $X'$\\
\end{flushleft}
\begin{algorithmic}

\STATE Compute the othonormal basis as \\$w'$ = qr-decomposition ($w$)  \# {\color{red} dim = ($h$, 3)}
\FOR{all samples}
\STATE Obtain the projected features on new basis via dot product: proj(X) = np.dot($X$, $w'$) \# {\color{red} dim = ($*$, 3)}
\STATE Dimension reduction from 3-D to 2-D via PCA(proj(X)) \# {\color{red} dim = ($*$, 2)}
\ENDFOR \\

{\bf \small RETURN} 2-D features: PCA(proj(X))
\end{algorithmic}
\label{sup-algorithm}
\end{algorithm*}

\begin{algorithm*}[h]
\caption{NumPy-style pseudo-code of the visualization algorithm} 

%%%%%%%%%%%%%%%%%%%%%%%%%%%%%%%%%%%
%%%%%%%%%%%%%%%%%%%%%%%%%%%%%%%%%%%
%%%%%%%%%%%%%%%%%%%%%%%%%%%%%%%%%%%
\definecolor{codegreen}{rgb}{0,0.6,0}
\definecolor{codegray}{rgb}{0.5,0.5,0.5}
\definecolor{codepurple}{rgb}{0.58,0,0.82}
\definecolor{backcolour}{rgb}{0.95,0.95,0.92}

\lstdefinestyle{mystyle}{
    backgroundcolor=\color{backcolour},   
    commentstyle=\color{codegreen},
    keywordstyle=\color{magenta},
    numberstyle=\tiny\color{codegray},
    stringstyle=\color{codepurple},
    basicstyle=\ttfamily\footnotesize,
    breakatwhitespace=false,         
    breaklines=true,                 
    captionpos=b,                    
    keepspaces=true,                 
    numbers=left,                    
    numbersep=5pt,                  
    showspaces=false,                
    showstringspaces=false,
    showtabs=false,                  
    tabsize=2
}
\lstset{style=mystyle}

\begin{lstlisting}[language=Python]
# Inputs
# weights_path: weights path of the final layer of your trained model
# feature_path: feature path of the penultimate layer high dimension features extracted by your trained model

# Outputs
# 2-D features of each class

# ------------------------------------------------------------------- #
# Step 0. Init settings and select the class to visualize
CLASSES = ['miniature_poodle', 'standard_poodle', 'submarine']
color = ['r', 'g', 'b']
model = 'resnet18'  # the student model

# Step 1. Compute the orthonormal basis
weights = np.load(weights_path) # load the final layer weights
basis, _ = np.linalg.qr(weights.T)  # dim=(*, 3)

# Step 2. Load the extracted features
num_sample = 150  # We sample 150 images per class
output_feature = np.load(feature_path)

# Step 3. Project the high dimension features to the new 3-D subspace
output_project = np.dot(output_feature, basis)

# Step 4. Dimension reduction from 3-D to 2-D using PCA
pca = PCA(n_components=2)
pca.fit(output_project)
output_array = pca.transform(output_project)

# Step 5. Plot the features on a 2-D plane
for i, subclass in enumerate(CLASSES):
    plt.scatter(output_array[i * num_sample:(i + 1) * num_sample, 0],
                output_array[i * num_sample:(i + 1) * num_sample, 1],
                c=color[i], label=subclass)
\end{lstlisting}

% \end{algorithmic}
\label{sup-algorithm-np}
\end{algorithm*}
%% -------------- End of Visualization algorithm ---------------

%% -------------- Semantically Similar and Dissimilar classes---------------
% \clearpage
\section{Semantically similar / dissimilar classes}
\label{sup-sec:semantic-classes}
\setcounter{figure}{0} 
\setcounter{table}{0} 

Given a target class $\pi$, let the set of semantically similar and dissimilar classes be $S_{1,}, S_{2}$ respectively. In this section, we discuss two important methods for identifying $S_1, S_2$ for the target class $\pi$.

\subsection{Method 1: Using standard, pre-defined ImageNet knowledge graph as a prior}
\label{sub-sec:sup-semantics-method1}
We use ImageNet hierarchy derived from WordNet \citep{Fellbaum1998} to select semantically similar classes and semantically dissimilar classes to quantify systematic diffusion. 
WordNet \citep{Fellbaum1998} is a laboriously hand-coded lexical database linking words into semantic relations including synonyms, hyponyms, and meronyms 
\footnote{\url{https://en.wikipedia.org/wiki/WordNet}}. 
Do note that ImageNet is organized using WordNet hierarchy. 
\href{https://observablehq.com/@mbostock/imagenet-hierarchy}{A web browser version of the ImageNet hierarchy can be accessed at this link (You can click any node to browse images that correspond to the associated synset)}

We use this ImagNet hierarchy to select semantically similar classes and semantically dissimilar classes for the target class $\pi$. 
This way, we ensure the selection of semantically similar classes ($S_1$) and semantically dissimilar classes ($S_2$) is based on a strong prior (knowledge graph) to support our main finding.

\subsection{Method 2: Using distance in the feature space to quantitatively define semantically similar / dissimilar classes}
\label{sub-sec:sup-semantics-method2}
This method is a quantitative approach for defining semantically similar / dissimilar classes.
Specifically, we consider the official ResNet-50 model trained on ImageNet-1K (classification). We use the validation set of ImageNet-1K and extract the penultimate layer representations for all the samples. 
For each class, we consider the centroid of the penultimate layer representations as the class prototype and calculate the centroid-centroid distance between all the classes (This will give a symmetric matrix of 1000 x 1000). 

\textit{For selecting $S_1$:} Next, for the target class $\pi$, we identify the \textit{closest 1\%} of classes (10 out of 999 classes) using the centroid-centroid distances.
These would be the semantically similar classes to the target class as they have the smallest distances to the centroid of the target class.

\textit{For selecting $S_2$:} Next, for the target class $\pi$,  we identify the \textit{distant 90\%} of classes (900 out of 999 classes) using the centroid-centroid distances discussed above. 
These would be the semantically dissimilar classes to the target class as their centroids lie much far away from the centroid of the target class. 

\textbf{Consistency measurements between the 2 methods:} 
Let the semantically similar and dissimilar classes identified using method 1 be $S_{1, qualitative}, S_{2, qualitative}$ respectively.
Let the semantically similar and dissimilar classes identified using method 1 be
$S_{1, quantitative}, S_{2, quantitative}$ respectively.
In this section, we measure the consistency between qualitative selection of $S_{1, qualitative}, S_{2, qualitative}$ (method 1) and the quantitative definition of $S_{1, quantitative}, S_{2, quantitative}$ (method 2). This consistency measurements are shown for all the target classes in the Table \ref{table:sup-consistency-measurements}. As one can clearly observe both method 1 and method 2 agree 85\% on average for semantically similar classes and 94\% on average for semantically dissimilar classes. 
Do note that we use pre-defined knowledge graph for ImageNet-1K as prior (method 1) to select the semantically similar / dissimilar classes for our $\eta$ computation in Table $\ref{table:eta}$.

\begin{table}[h]
    \centering
    \caption{
    Consistency measurements between using pre-defined knowledge graph for ImageNet-1K as prior vs. feature space distance method for identifying semantically similar / dissimilar classes.
    This table shows the agreement between these 2 methods in identifying semantically similar / dissimilar classes. 
    Each row indicates the agreement between the 2 methods with respect to the target class. An agreement value of 1.000 indicates a perfect agreement between the 2 methods. 
    As we can clearly observe on average both methods agree 85\% for semantically similar classes and 94\% for semantically dissimilar classes.
    This can suggest that we can leverage on either one of the methods to select the semantically similar / dissimilar classes for our analysis on systematic diffusion.
    Do note that we use pre-defined knowledge graph for ImageNet-1K as prior (method 1) to select the semantically similar / dissimilar classes for our $\eta$ computation in Table $\ref{table:eta}$.
    }
        
    \begin{adjustbox}{width=0.95\columnwidth,center}
        \begin{tabular}{lcc}\toprule
        Target class &
        $\frac{ S_{1, qualitative} \cap  S_{1, quantitative} }{\Vert S_{1, qualitative} \Vert}$ &
        $\frac{ S_{2, qualitative} \cap  S_{2, quantitative} }{\Vert S_{2, qualitative} \Vert}$\\ \toprule
        Chesapeake Bay retriever &1.000 &0.950 \\ \midrule
        curly-coated retriever &0.750 &0.950 \\ \midrule
        flat-coated retriever &1.000 &1.000 \\ \midrule
        golden retriever &0.500 &1.000 \\ \midrule
        Labrador retriever &0.750 &1.000 \\ \midrule
        thunder\_snake &1.000 &0.900 \\ \midrule
        ringneck\_snake &1.000 &0.900 \\ \midrule
        hognose\_snake &0.500 &0.900 \\ \midrule
        water\_snake &1.000 &0.900 \\ \midrule
        king\_snake &1.000 &0.900 \\ \midrule
        \textbf{Average} &\textbf{0.850} &\textbf{0.940} \\
        \bottomrule
        \end{tabular}

    \end{adjustbox}
    
\label{table:sup-consistency-measurements}
\end{table}
%% -------------- End of Semantically Similar and Dissimilar classes---------------

%% -------------- Can smoothness determine the KD performance?--------------
% \pagebreak
% \quad
% \newpage
% \quad
% \pagebreak
\section{Case study: Smoothness of targets are insufficient to determine KD performance. Systematic diffusion is critical.}
\label{sup-sec:smoothness}
\setcounter{figure}{0} 
\setcounter{table}{0} 

An interesting perspective is whether the degree of smoothness of targets produced by an LS-trained teacher can determine the KD performance (of the student). 
We acknowledge that smoothness of targets produced by the teacher at different temperatures is important. 
However, we quantitatively show that the degree of smoothness cannot adequately explain the KD performance in the presence of an LS-trained teacher. 
More specifically, we show that the KD performance in the presence of LS-trained teachers can be explained by our discovered systematic diffusion and not directly using the degree of smoothness. 
The detailed study is discussed below.

\textbf{Our view:} The degree of smoothness of targets is rather unable to explain the performance of KD. We show this using 3 comprehensive case studies comprising 7 counterexamples. 

\textbf{Measuring smoothness of targets:} To perform a quantitative study to support our view, we measure the smoothness of the targets produced by the teacher. 
The target produced for every training sample by the teacher for KD is a discrete probability distribution. 
To measure the smoothness of this target, we can use entropy which is a very popular method. Entropy of a discrete probability distribution with $N$ classes can be indicated by $H(p) = \sum_{i}^{N} - p_i ln(p_i)$ where $p_i$ indicates the probability assigned to the $i^{th}$ class. 
The maximum entropy/smoothness will be equal to $H_{max}(p) = ln(N)$ which corresponds to the uniform probability distribution over all classes. 
\textit{The key idea here is higher the entropy, smoother the target. }
We measure the average entropy for the training set (since this is the set used for distillation) to approximate the smoothness of the targets. Do note that the average entropy is measured using the targets produced by the teacher at different $T$.

Table \ref{table:sup-entropy} shows the average entropy/  smoothness of the targets for the ResNet-50 teachers used in our CUB200-2011 experiments. Higher entropy indicates that the targets are over-smoothed. Do note that the maximum average entropy for CUB200-2011 \citep{WahCUB_200_2011} is $ln(200) \approx 5.298$.

\subsection{Case study at lower $T$ with same degree of smoothness}
\label{sub-sec:sup-case-study-1}
Consider a lower $T$. 

As shown in Table \ref{table:sup-entropy}, the entropy / smoothness of targets produced by LS-trained teacher ($\alpha=0.1$) at $T=1$ is approximately equal to the entropy/ smoothness of targets produced by normally-trained teacher ($\alpha=0.0$) at $T=1.481375$. 
If smoothness of targets can determine the KD performance, then we expect comparable performances in both the instances above as they have the same degree of smoothness. 

But using 2 counterexamples shown in Table \ref{table:sup-case-study-1}, we show that even at the same degree of smoothness,  distilling from LS-trained teachers produces better students compared to distilling from normally-trained teachers at lower $T$ due to lower degree of systematic diffusion (LS and KD are compatible). 
Through these counterexamples we show that whether or not LS was used during training of teacher is very important in determining the performance of distillation even at the same degree of smoothness, thereby showing that the degree of smoothness is insufficient/ unreliable in determining the performance of distillation.

%% ======Case study 2===================
\subsection{Case study at moderately higher $T$ with same degree of smoothness}
\label{sub-sec:sup-case-study-2}
Consider a moderately higher $T$. 

As shown in Table \ref{table:sup-entropy}, the entropy / smoothness of targets produced by LS-trained teacher ($\alpha=0.1$) at $T=3$ is approximately equal to the entropy/ smoothness of targets produced by normally-trained teacher ($\alpha=0.0$) at $T=5.638$. If the smoothness is the most important factor, then we expect comparable performances in both the instances above as they have the same degree of smoothness. 

But using 2 counterexamples shown in Table \ref{table:sup-case-study-2}, we show that even at the same degree of smoothness,  distilling from LS-trained teachers produces poorer students compared to distilling from normally-trained teachers at moderately higher $T$ due to increased degree of systematic diffusion (LS and KD are incompatible). Through these counterexamples we show that whether LS was used during training of teacher or not is very important in determining the performance of distillation even at the same degree of smoothness, thereby showing that the degree of smoothness is insufficient/ unreliable in determining the performance of distillation.

%% ======Case study 3===================
\subsection{Case study at extremely high $T$ with same degree of smoothness}
\label{sub-sec:sup-case-study-3}
Consider a very high $T$. 

As shown in Table \ref{table:sup-entropy}, the entropy / smoothness of targets produced by LS-trained teacher ($\alpha=0.1$) at $T=64$ is approximately equal to the entropy/ smoothness of targets produced by normally-trained teacher ($\alpha=0.0$) at $T=64$ since at very high $T$ both these models produce a probability distribution that is very close to the uniform distribution. If the smoothness is the most important factor, then we expect comparable performances in both the instances above as they have the same degree of smoothness. 

But using 3 counterexamples shown in Table \ref{table:sup-case-study-3}, we show that even at the same degree of smoothness,  distilling from LS-trained teachers produces poorer students compared to distilling from normally-trained teachers at extremely higher $T$ due to extreme degree of systematic diffusion (LS and KD are incompatible). Through these counterexamples we show that whether LS was used during training of teacher or not is very important in determining the performance of distillation even at the same degree of smoothness, thereby showing that the degree of smoothness is insufficient/ unreliable in determining the performance of distillation.

\textbf{Conclusion regarding smoothness: }Through these 3 quantitative case studies comprising of 7 counterexamples, we show that whether or not LS was used during training of teacher is very important in determining the performance of distillation even at the same degree of smoothness, thereby showing that the degree of smoothness is insufficient/ unreliable in determining the performance of distillation. 

Another way to intuitively think about this is that smoothness of targets can be characterized using the probability output of the teacher at different temperatures. \textit{But systematic diffusion is a phenomenon happening exclusively in the student. 
This is precisely the reason why we quantify the degree of systematic diffusion using penultimate layer representations of the student, as these student representations are more indicative of the resulting student performance.}
That is, in all our 34 experiments, increased systematic diffusion definitely indicates lower performance of students whereas the degree of smoothness of targets does not give reliable insights as shown in the case studies \ref{sub-sec:sup-case-study-1}, \ref{sub-sec:sup-case-study-2}, \ref{sub-sec:sup-case-study-3}.
%% -------------- End of Can smoothness determine the KD performance?--------------

\begin{table*}
    \centering
    \caption{
    This table shows the degree of smoothness as measured by average entropy using the training set of CUB200-2011 at different temperatures for normally trained ResNet-50 teacher and LS-trained ResNet-50 teacher. 
    Do note that this analysis is done using CUB200-2011. 
    We make important observations regarding the smoothness of the targets produced by LS-trained teachers and teachers training without LS.
    (1) As one can observe, at $T=1$, LS-trained teacher produces smoother targets compared to the normal teacher.
    (2) As $T$ increases, the targets become smoother. At moderate levels of $T$ (See $T=2, 3$), the LS-trained teacher will produce over smoothed targets compared to the normal teacher.
    (3) At very high $T$ (See $T=64$), both LS-trained teacher and normal teacher will have almost the same amount of smoothness (almost closer to maximum entropy) as they produce a probability distribution that is very close to the uniform distribution.
    We particularly identify pairs of specific temperatures where the entropy/ smoothness of normally-trained teacher is approximately equal to a configuration of LS-trained teacher in the table. These pairs are in\textbf{ bold }. I..e: The entropy / smoothness of targets produced by LS-trained teacher ($\alpha=0.1$) at $T=1$ is approximately equal to the entropy/ smoothness of targets produced by normally-trained teacher ($\alpha=0.0$) at $T=1.481375$ which is $\approx$ 0.888. 
    }
        
    \begin{adjustbox}{width=0.85\textwidth,center}
        \begin{tabular}{ccc}\toprule
        \textbf{CUB200-2011 Training Set: Average Entropy of the targets from ResNet-50 teacher} &$\alpha=0.0$ &$\alpha=0.1$ \\ \toprule
        $T=1$ &0.184 &\textbf{0.888} \\ \midrule
        $T=1.481375$ &\textbf{0.888} &3.225 \\ \midrule
        $T=2$ &2.246 &4.550 \\ \midrule
        $T=3$ &4.160 &\textbf{5.118} \\ \midrule
        $T=5.638$ &\textbf{5.118} &5.269 \\ \midrule
        $T=64$ &\textbf{5.298} &\textbf{5.298} \\ 
        \bottomrule
        \end{tabular}

    \end{adjustbox}
    
\label{table:sup-entropy}
\end{table*}

\begin{table*}
    \centering
    \caption{
    Results of case study at lower $T$ with same degree of smoothness.
    In Counterexample \#1, Teacher is ResNet-50, Student is ResNet-50. Two $\alpha /T$ configurations have been identified such that average entropy of the teachers’ output are the same (0.888). We clearly observe different performances for Student.
    Similarly, in Counterexample \#2, Teacher is ResNet-50, Student is ResNet-18 and we clearly observe different performances for Student. 
    For each counterexample, the higher KD performance is in \textbf{bold}.
    Through these 2 counterexamples, we show that even at the same degree of smoothness,  distilling from LS-trained teachers produces better students compared to distilling from normally-trained teachers at lower $T$ due to lower degree of systematic diffusion (LS and KD are compatible). 
    }
        
    \begin{adjustbox}{width=0.88\textwidth,center}
        \begin{tabular}{l|c|c|c|c}\toprule
        Counterexample &Student &$\alpha / T$ &Average Entropy &KD performance: Top1/Top5 \\ \toprule
        \multirow{2}{*}{\#1} &ResNet-50 &$\alpha=0.1 / T=1.0$ &0.888 &\textbf{83.742 / 96.778} \\ \cmidrule{2-5}
        &ResNet-50 &$\alpha=0.0 / T=1.481375$ &0.888 &82.603 / 96.496 \\ \midrule
        \multirow{2}{*}{\#2} &ResNet-18 &$\alpha=0.1 / T=1.0$ &0.888 &\textbf{80.946 / 95.312} \\ \cmidrule{2-5}
        &ResNet-18 &$\alpha=0.0 / T=1.481375$ &0.888 &80.808 / 95.547 \\
        \bottomrule
        \end{tabular}

    \end{adjustbox}
    
\label{table:sup-case-study-1}
\end{table*}

\begin{table*}
    \centering
    \caption{
    Results of case study at moderately higher $T$ with same degree of smoothness. 
    In Counterexample \#3, Teacher is ResNet-50, Student is ResNet-18. Two $\alpha /T$ configurations have been identified such that average entropy of the teachers’ output are the same (5.188). We clearly observe different performances for Student.
    Similarly, in Counterexample \#4, Teacher is ResNet-50, Student is MobileNetV2 and we clearly observe different performances for Student. 
    For each counterexample, the higher KD performance is in \textbf{bold}.
    Through these 2 counterexamples, we show that even at the same degree of smoothness, distilling from LS-trained teachers produces poorer students compared to distilling from normally-trained teachers. This is due to increased degree of systematic diffusion as $T$ increases in the presence of LS-trained teachers, thereby producing poor students (LS and KD are incompatible).
    }
        
    \begin{adjustbox}{width=0.88\textwidth,center}
        \begin{tabular}{l|c|c|c|c}\toprule
        Counterexample &Student &$\alpha / T$ &Average Entropy &Student performance: Top1/Top5 \\ \toprule
        \multirow{2}{*}{\#3} &ResNet-18 &$\alpha=0.1 / T=3.0$ &5.118 &78.196 / 95.213 \\ \cmidrule{2-5}
        &ResNet-18 &$\alpha=0.0 / T=5.638$ &5.118 &\textbf{78.719 / 95.478} \\ \midrule
        \multirow{2}{*}{\#4} &MobileNetV2 &$\alpha=0.1 / T=3.0$ &5.118 &78.961 / 95.306 \\ \cmidrule{2-5}
        &MobileNetV2 &$\alpha=0.0 / T=5.638$ &5.118 &\textbf{79.341 / 95.461} \\
        \bottomrule
        \end{tabular}

    \end{adjustbox}
    
\label{table:sup-case-study-2}
\end{table*}

\begin{table*}
    \centering
    \caption{
    Results of case study at extremely high $T$ with same degree of smoothness. 
    In Counterexample \#5, Teacher is ResNet-50, Student is ResNet-18. Two $\alpha /T$ configurations have been identified such that average entropy of the teachers’ output are the same (5.298). We clearly observe different performances for Student.
    Similarly, in Counterexample \#6, Teacher is ResNet-50, Student is ResNet-50 and we clearly observe different performances for Student. 
    In Counterexample \#7, Teacher is ResNet-50, Student is MobileNetV2 and we clearly observe different performances for Student.
    For each counterexample, the higher KD performance is in \textbf{bold}.
    Through these 3 counterexamples, we show that even at the same degree of smoothness, distilling from LS-trained teachers produces extremely poorer students compared to distilling from normally-trained teachers. This is due to extreme degree of systematic diffusion at very high $T$ in the presence of LS-trained teachers, thereby producing poor students (LS and KD are incompatible).
    }
        
    \begin{adjustbox}{width=0.85\textwidth,center}
        \begin{tabular}{l|c|c|c|c}\toprule
        Counterexample &Student &$\alpha / T$ &Average Entropy &Student performance: Top1/Top5 \\ \toprule
        \multirow{2}{*}{\#5} &ResNet-18 &$\alpha=0.1 / T=64$ &5.298 &67.161 / 93.062 \\ \cmidrule{2-5}
        &ResNet-18 &$\alpha=0.0 / T=64$ &5.298 &\textbf{73.611 / 94.529} \\ \midrule
        \multirow{2}{*}{\#6} &ResNet-50 &$\alpha=0.1 / T=64$ &5.298 &77.206 / 95.812 \\ \cmidrule{2-5}
        &ResNet-50 &$\alpha=0.0 / T=64$ &5.298 &\textbf{79.784 / 95.927} \\ \midrule
        \multirow{2}{*}{\#7} &MobileNetV2 &$\alpha=0.1 / T=64$ &5.298 &70.435 / 93.494 \\ \cmidrule{2-5}
        &MobileNetV2 &$\alpha=0.0 / T=64$ &5.298 &\textbf{75.441 / 94.702} \\
        \bottomrule
        \end{tabular}

    \end{adjustbox}
    
\label{table:sup-case-study-3}
\end{table*}

%% -------------- Classwise accuracies--------------
% \clearpage
\section{Class-wise accuracy for target classes }
\label{sup-sec:class-wise-accuracy}
\setcounter{figure}{0} 
\setcounter{table}{0}

This section contains class-wise accuracy for all the target classes used in the paper. 

\begin{table}[t]
    \centering
    \caption{
    The table shows the class-wise accuracies for the 3 classes used in Figure\ref{fig:main} (penultimate layer visualization). 
    As one can observe, in the presence of an LS-trained teacher, KD at higher temperatures causes systematic diffusion thereby rendering KD ineffective. 
    We can see this for most classes at increased temperatures shown below. That is, in the presence of an LS-trained teacher as we increase the temperature from $T=1$, the accuracies for most of these classes drop due to systematic diffusion. 
    This can be seen in both training and validation sets. Do note that since the validation set contains only 50 samples per class, class wise validation accuracies may not be statistically reliable and contain outlier points, and we suggest observing the general trend as shown by the average for the set.
    }
        
    \begin{adjustbox}{width=0.99\columnwidth,center}
        \begin{tabular}{l|cc|cc|cc}\toprule
        Set A (Figure\ref{fig:main}) &\multicolumn{2}{c}{$T=1$} &\multicolumn{2}{c}{$T=2$} &\multicolumn{2}{c}{$T=3$} \\ \toprule
        \textbf{} &Train &Val &Train &Val &Train &Val \\ \midrule
        miniature\_poodle &58.077 &46.000 &47.462 &46.000 &49.846 &34.000 \\ \midrule
        standard\_poodle &72.077 &80.000 &65.462 &76.000 &61.846 &74.000 \\ \midrule
        submarine &89.692 &68.000 &85.077 &64.000 &82.000 &54.000 \\ \midrule
        \textbf{Average} &\textbf{73.282} &\textbf{64.667} &\textbf{66.000} &\textbf{62.000} &\textbf{64.564} &\textbf{54.000} \\
        \bottomrule
        \end{tabular}

    \end{adjustbox}
    
\label{table:sup-class-wise-seta}
\end{table}

Given that we use the training set for distillation, let us consider both the training set and the validation set for this analysis. 
There are 1300 training and 50 validation samples for each class in ImageNet-1k. 
We use an exhaustive list of $T$ values for this analysis, $T=1, T=2, T=3$, and use the exact LS-trained teacher (ResNet-50, $\alpha=0.1$) reported in Table \ref{table:results-imagenet-cub}.
There are 13 target classes used: 3 classes for the visualization in Figure \ref{fig:main}, and 10 classes in Table \ref{table:eta}.  
We show the complete class wise accuracies for both the training and validation set at $T=1, T=2, T=3$. 
For each set we also compute the average accuracies to show the general trend to support our main findings. The results are shown in Tables \ref{table:sup-class-wise-seta}, \ref{table:sup-class-wise-setb} and \ref{table:sup-class-wise-setc}.
As one can observe in Tables \ref{table:sup-class-wise-seta}, \ref{table:sup-class-wise-setb}, \ref{table:sup-class-wise-setc} , in the presence of an LS-trained teacher, KD at higher temperatures causes systematic diffusion thereby rendering KD ineffective. 
We can see this for most classes at increased temperatures shown below. That is, in the presence of an LS-trained teacher as we increase the temperature from $T=1$, the accuracies for most of these classes drop due to systematic diffusion. 
This can be seen in both training and validation sets.

\begin{table}[t]
    \centering
    \caption{
    The table shows the class-wise accuracies for the 5 targets classes used in our systematic diffusion analysis ($\eta$ calculation as shown in \ref{table:eta_set1}).
    As one can observe, in the presence of an LS-trained teacher, KD at higher temperatures causes systematic diffusion thereby rendering KD ineffective. 
    We can see this for most classes at increased temperatures shown below. That is, in the presence of an LS-trained teacher as we increase the temperature from $T=1$, the accuracies for most of these classes drop due to systematic diffusion. 
    This can be seen in both training and validation sets. Do note that since the validation set contains only 50 samples per class, class wise validation accuracies may not be statistically reliable and contain outlier points, and we suggest observing the general trend as shown by the average for the set.
    }
        
    \begin{adjustbox}{width=0.99\columnwidth,center}
        \begin{tabular}{l|cc|cc|cc}\toprule
        Set B &\multicolumn{2}{c}{$T=1$} &\multicolumn{2}{c}{$T=2$} &\multicolumn{2}{c}{$T=3$} \\ \toprule
        \textbf{} &Train &Val &Train &Val &Train &Val \\ \midrule
        Chesapeake Bay retriever &86.308 &84.000 &80.846 &80.000 &78.846 &76.000 \\ \midrule
        curly-coated retriever &83.826 &76.000 &81.199 &82.000 &80.296 &74.000 \\ \midrule
        flat-coated retriever &82.538 &80.000 &79.154 &72.000 &79.462 &70.000 \\ \midrule
        golden retriever &81.154 &86.000 &75.615 &84.000 &76.000 &76.000 \\ \midrule
        Labrador retriever &70.692 &82.000 &62.692 &86.000 &58.385 &78.000 \\ \midrule
        \textbf{Average} &\textbf{80.900} &\textbf{81.600} &\textbf{75.900} &\textbf{80.800} &\textbf{74.600} &\textbf{74.800} \\
        \bottomrule
        \end{tabular}

    \end{adjustbox}
    
\label{table:sup-class-wise-setb}
\end{table}

\begin{table}
    \centering
    \caption{
    The table shows class-wise accuracies for the 5 target classes used in our systematic diffusion analysis ($\eta$ calculation as shown in Table \ref{table:eta_set2}).
    As one can observe, in the presence of an LS-trained teacher, KD at higher temperatures causes systematic diffusion thereby rendering KD ineffective. 
    We can see this for most classes at increased temperatures shown below. That is, in the presence of an LS-trained teacher as we increase the temperature from $T=1$, the accuracies for most of these classes drop due to systematic diffusion. 
    This can be seen in both training and validation sets. Do note that since the validation set contains only 50 samples per class, class wise validation accuracies may not be statistically reliable and contain outlier points, and we suggest observing the general trend as shown by the average for the set.
    }
        
    \begin{adjustbox}{width=0.99\columnwidth,center}
        \begin{tabular}{l|cc|cc|cc}\toprule
        Set B &\multicolumn{2}{c}{$T=1$} &\multicolumn{2}{c}{$T=2$} &\multicolumn{2}{c}{$T=3$} \\ \toprule
        \textbf{} &Train &Val &Train &Val &Train &Val \\ \midrule
        thunder\_snake &84.615 &78.000 &69.231 &68.000 &68.462 &66.000 \\ \midrule
        ringneck\_snake &70.000 &86.000 &78.923 &82.000 &77.538 &78.000 \\ \midrule
        hognose\_snake &76.692 &60.000 &60.154 &56.000 &52.000 &42.000 \\ \midrule
        water\_snake &86.154 &64.000 &67.385 &60.000 &68.385 &72.000 \\ \midrule
        king\_snake &58.077 &78.000 &80.385 &72.000 &79.692 &78.000 \\ \midrule
        \textbf{Average} &\textbf{75.110} &\textbf{73.200} &\textbf{71.220} &\textbf{67.600} &\textbf{69.220} &\textbf{67.200} \\
        \bottomrule
        \end{tabular}

    \end{adjustbox}
    
\label{table:sup-class-wise-setc}
\end{table}

% ==================== \alpha and T =====================
% \clearpage
\section{Additional Exploration of $\alpha$ and $T$}
\label{sup-sec:additional_alpha_T}
\setcounter{figure}{0} 
\setcounter{table}{0} 

\begin{table}
    \centering
    \caption{
    The table shows results of additional exploration of $\alpha$ and $T$. 
    CUB200-2011 dataset / MobileNetV2 setup is used for these experiments.
    }
        
    \begin{adjustbox}{width=0.99\columnwidth,center}
        \begin{tabular}{l|c|c|c|c}\toprule
         &\backslashbox{$T$}{$\alpha$} & $\alpha=0$ & $\alpha=0.1$ & $\alpha=0.2$ \\ \toprule
        Teacher : ResNet-50 &	- &	81.584 / 95.927	& 82.068 / 96.168 &	81.412 / 96.186 \\ \midrule
\multirow{4}{*}{Student : MobileNetV2} &	T=1	 & 81.144 / 95.677	& {\bf 81.731 / 95.754}	& {\bf 81.498 / 95.892} \\ \cmidrule{2-5}
 &	T=2	 & 81.895 / 95.858 &	80.609 / 95.470 &	79.997 / 95.599 \\ \cmidrule{2-5}
 &	T=3	& 81.257 / 95.677 &	78.961 / 95.306	 & 76.959 / 95.202 \\ \cmidrule{2-5}
 &	T=64 &	75.441 / 94.702 &	70.435 / 93.494 &	63.738 / 91.992 \\
        \bottomrule
        \end{tabular}

    \end{adjustbox}
    
\label{table:additional_alpha_T}
\end{table}

Given that label smoothing was originally formulated as a regularization strategy to alleviate models' overconfidence, most works spanning different learning problems use a smaller $\alpha = 0.1$, including work closely related to our study. The intuition is that a larger $\alpha$ can introduce too much regularization that may subsequently hurt the model performance.

To show this, here we conduct additional experiments using larger $\alpha$ ($\alpha = 0.2$) for compact student distillation. We use CUB200-2011 dataset for these experiments.

The results are shown
in Table 
\ref{table:additional_alpha_T}.
These additional results further support our findings on systematic diffusion.

In particular, we can make two important observations here: (i) 
larger $\alpha$ ($\alpha =0.2$)
results in a weaker ResNet-50 teacher. We emphasize that it is reasonable to expect such behaviour, and this suggests  why most works  use $\alpha =0.1$ as in our main experiments. (ii)
As one can clearly observe, with $\alpha =0.2$, KD at higher $T$  causes systematic diffusion, thereby rendering KD substantially ineffective.

These experiments further support our main finding, and we emphasize that our findings can be generalized to larger values of $\alpha$ ($\alpha =0.2$).

%========================
% \clearpage
\section{Alternative characterization of cluster distance}
\label{sup-sec:alternative_characterization}
\setcounter{figure}{0} 
\setcounter{table}{0} 

\begin{table}
    \centering
    \caption{Results of using alternative distance, i.e., pairwise distance, to define the 
    diffusion index
$\eta_{pairwise}$. Our findings on systematic diffusion are consistent with 
using alternative distance characterization.
    }
        
    \begin{adjustbox}{width=0.93\columnwidth,center}
        \begin{tabular}{l|c|c|c|c}\toprule
         &Train: $S_1$ & Train: $S_2$ & Val: $S_1$ & Val: $S_2$ \\ \toprule
Chesapeake Bay retriever &	-2.532	& 1.025 &	-2.919	& 1.154 \\ \midrule
curly-coated retriever &	-2.359 & 	1.208 &	-3.068 &	1.354 \\ \midrule
flat-coated retriever &	-3.201 & 	1.183 &	-3.643 &	1.237 \\ \midrule
golden retriever &	-2.307 & 	0.895 &	-2.994	& 1.038 \\ \midrule
Labrador retriever	& -3.586	& 1.089 & 	-4.337 &	1.355 \\ \midrule
thunder\_snake	& -5.438 & 	1.642 &	-6.419	& 1.939 \\ \midrule
ringneck\_snake	& -5.680 &	1.814 &	-5.914	& 1.775 \\ \midrule
hognose\_snake	& -5.327 &	1.742 &	-5.393	& 1.707 \\ \midrule
water\_snake &	-5.266 & 	1.672 &	-5.301 & 	1.640 \\ \midrule
king\_snake	& -5.454 &	1.941 &	-5.783	& 1.998 \\

        \bottomrule
        \end{tabular}

    \end{adjustbox}
    
\label{table:alternative_characterization}
\end{table}

Here we discuss an 
 alternative characterization of cluster distance  based on pairwise distances.
 
While our proposed $\eta$
(Table     \ref{table:eta})
to use centroids to characterise distance between clusters should be very robust, here we discuss an alternative. 

In this alternative, we propose to replace centroid-centroid distance with {\em average pairwise distance} between the projected penultimate layer representations. 
Note that this alternative is more computationally expensive.

We perform additional experiments using this alternative pairwise distance metric. We show that  diffusion index based on this alternative distance,
$\eta_{pairwise}$,
for all the 10 target classes used in the paper with this pairwise distance below (see Table \ref{table:alternative_characterization}).

As one can clearly observe, using this alternative (pairwise distances) we obtain consistent findings
for all 10 target classes as that in the paper Table 
    \ref{table:eta}: 
negative 
$\eta_{pairwise}$ for $S_1$,
positive
$\eta_{pairwise}$ for $S_2$.

%=====================
% \clearpage

\section{Sample images}
\label{sup-sec:sample_images}
\setcounter{figure}{0} 
\setcounter{table}{0} 

\begin{figure}
\begin{adjustbox}{width=0.95\columnwidth,center}
\begin{tabular}{c}
    \includegraphics[width=0.9\columnwidth]{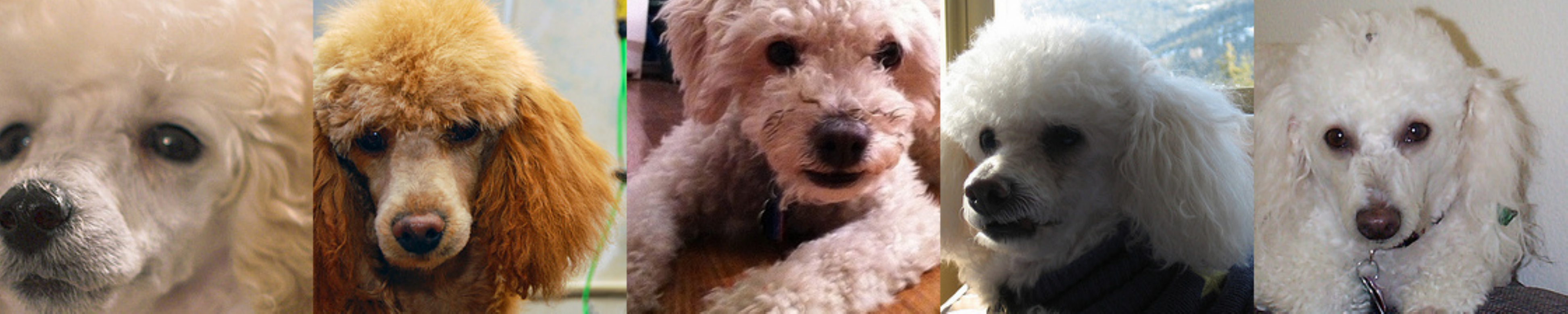} \\
    \includegraphics[width=0.9\columnwidth]{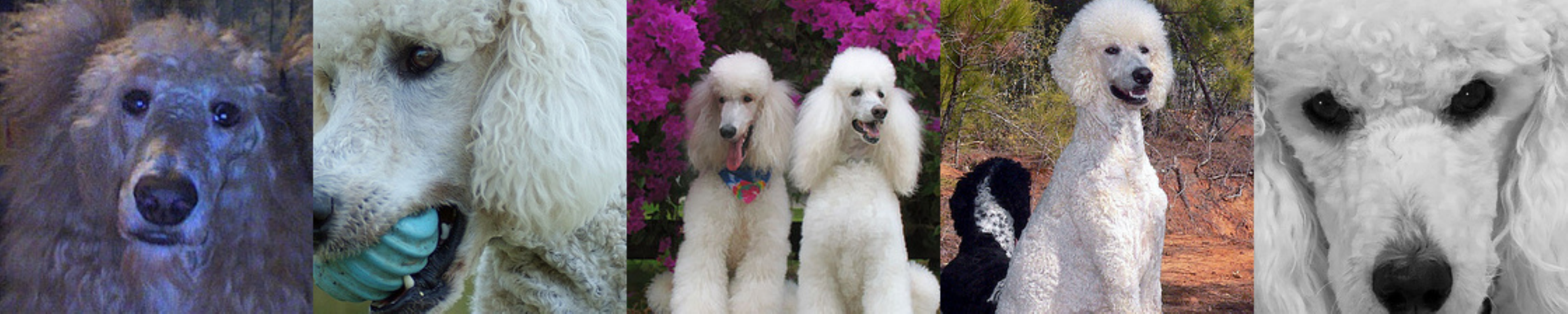} \\
    \includegraphics[width=0.9\columnwidth]{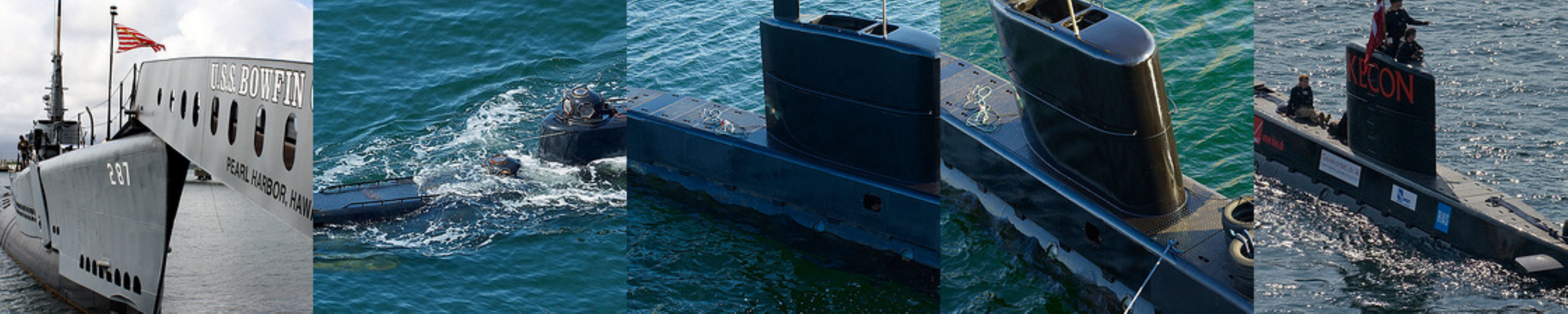}
\end{tabular}
\end{adjustbox}
\caption{
We show 5 samples of miniature\_poodle, standard\_poodle and submarine classes in top, middle and bottom rows respectively. 
These samples are obtained from the ImageNet-1K validation set \citep{imagenet_cvpr09}. 
As one can observe miniature\_poodle and standard\_poodle are semantically similar (They belong to the same category poodle).
On the other hand submarine class is semantically dissimilar to both  miniature\_poodle and standard\_poodle classes.
We can clearly observe the systematic diffusion at increased $T$ in the presence of an LS-trained teacher for the semantically similar classes from the penultimate layer visualizations shown in Figures \ref{fig:main}, \ref{sup-fig:r50-imagenet} and \ref{sup-fig:efb0-imagenet}.
}
\label{sup-fig:samples_imagenet}
\end{figure}

\begin{figure}
\begin{adjustbox}{width=0.95\columnwidth,center}
\begin{tabular}{c}
    \includegraphics[width=0.9\columnwidth]{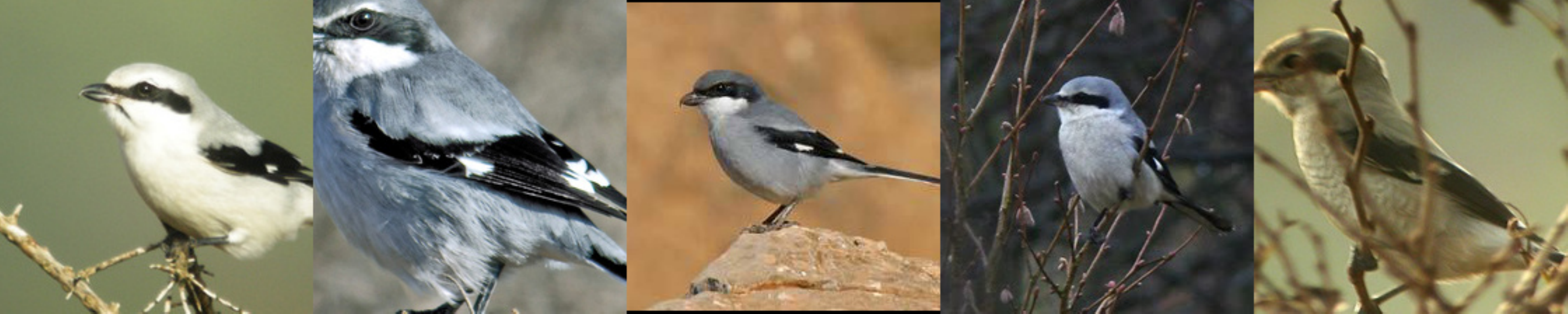} \\
    \includegraphics[width=0.9\columnwidth]{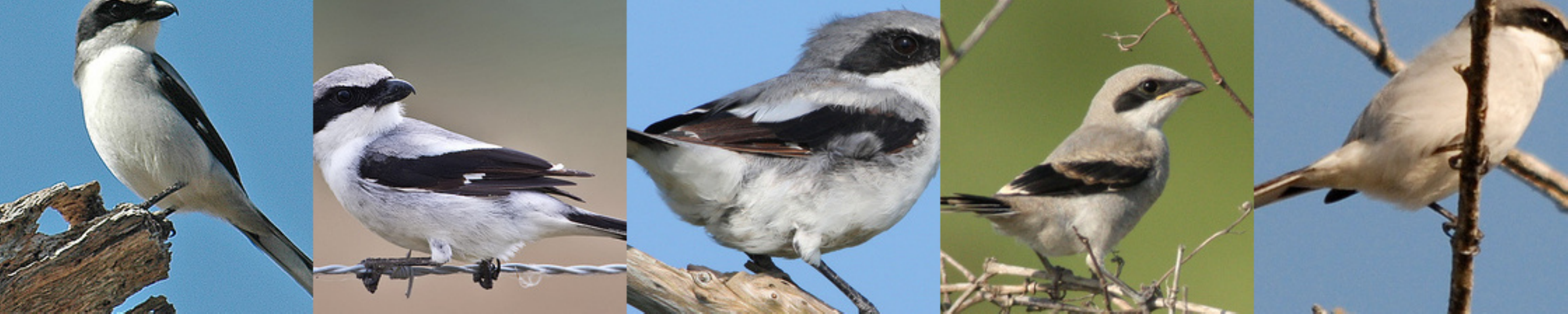} \\
    \includegraphics[width=0.9\columnwidth]{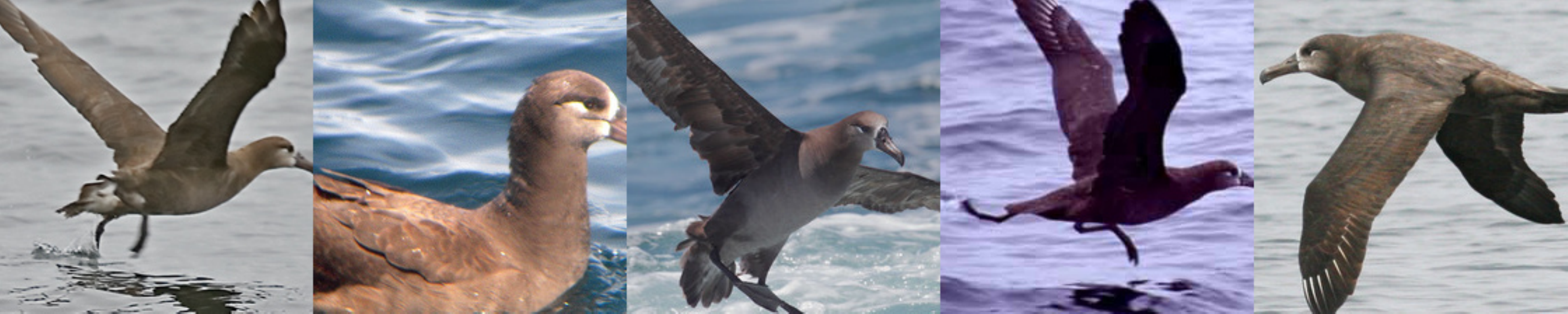}
\end{tabular}
\end{adjustbox}
\caption{
We show 5 samples of Great\_grey\_shrike, loggerhead\_shrike and black\_footed\_albatross classes in top, middle and bottom rows respectively. 
These samples are obtained from the CUB200-2011 validation set \citep{WahCUB_200_2011}. 
As one can observe Great\_grey\_shrike and loggerhead\_shrike are semantically similar (They belong to the same category shrike).
On the other hand black\_footed\_albatross class is semantically dissimilar to both  Great\_grey\_shrike and loggerhead\_shrike classes.
We can clearly observe the systematic diffusion at increased $T$ in the presence of an LS-trained teacher for the semantically similar classes from the penultimate layer visualizations shown in Figures \ref{sup-fig:r18-cub}, \ref{sup-fig:r50-cub} and \ref{sup-fig:convnext-cub}.
}
\label{sup-fig:samples_cub}
\end{figure}

In this section, we include samples images from 3 different classes used in the penultimate layer visualizations for ImageNet-1K and CUB200-2011 experiments.
Refer to Figures \ref{sup-fig:samples_imagenet} and \ref{sup-fig:samples_cub} for ImageNet-1K and CUB200-2011 samples respectively.

\end{document}